\pdfoutput=1

\documentclass[11pt]{article}

\usepackage[]{acl}

\usepackage{times}
\usepackage{latexsym}

\usepackage[T1]{fontenc}

\usepackage[utf8]{inputenc}

\usepackage{microtype}
\usepackage{adjustbox}
\usepackage{booktabs}
\usepackage{multirow}
\usepackage{amsmath}
\usepackage{amssymb}
\usepackage{amsthm}
\usepackage{enumitem}
\usepackage{bm}
\usepackage{bbm}
\usepackage{algorithm}
\usepackage{algcompatible}
\usepackage{algpseudocode}
\usepackage{xcolor}
\usepackage{xspace}
\usepackage{graphicx}
\usepackage{subfig}

\def\figref#1{Fig.~\ref{#1}}
\def\secref#1{Sec.~\ref{#1}}
\def\appendixref#1{Appendix~\ref{#1}}
\def\tabref#1{Table~\ref{#1}}
\def\eqnref#1{Eqn.~\ref{#1}}

\def\theoremref#1{Theorem.~\ref{#1}}

\newcommand{\modelname}{\textit{SQ-Transformer}\xspace}
\newcommand{\modelnamenospace}{\textit{SQ-Transformer}}

\newcommand{\addjump}{\textsc{AddJump}\xspace}

\newcommand{\aroundright}{\textsc{AroundRight}\xspace}

\newlist{Properties}{enumerate}{2}
\setlist[Properties]{label=Property \arabic*, font=\textbf, itemindent=*, align=left}

\newlist{Definitions}{enumerate}{2}
\setlist[Definitions]{label=Definition \arabic*, font=\textbf, itemindent=*, align=left}

\newlist{Conditions}{enumerate}{2}
\setlist[Conditions]{label=Hypothesis \arabic*, font=\textbf, itemindent=*, align=left}

\newlist{Conditions2}{enumerate}{2}
\setlist[Conditions2]{label=2.\Alph*, font=\textbf, itemindent=*, align=left}

\newtheorem{theorem}{Theorem}

\title{Inducing Systematicity in Transformers by \\Attending to Structurally Quantized Embeddings}

\author{Yichen Jiang \ \ \ \ \ \ \ \ Xiang Zhou \ \ \ \ \ \ \ \ Mohit Bansal \\
UNC Chapel Hill \\
  \texttt{\{yichenj, xzh, mbansal\}@cs.unc.edu} \\
}

\begin{document}
\maketitle
\begin{abstract}

Transformers generalize to novel compositions of structures and entities after being trained on a complex dataset, 
but easily overfit on datasets of insufficient complexity. 
We observe that when the training set is sufficiently complex, the model encodes sentences that have a common syntactic structure using a systematic attention pattern.
Inspired by this observation, we propose \modelname (\textbf{S}tructurally \textbf{Q}uantized) that explicitly encourages systematicity in the embeddings and attention layers, even with a training set of low complexity.
At the embedding level, we introduce Structure-oriented Vector Quantization (SoVQ) to cluster word embeddings into several classes of structurally equivalent entities.
At the attention level, we devise the Systematic Attention Layer (SAL) and an alternative, Systematically Regularized Layer (SRL) that operate on the quantized word embeddings so that sentences of the same structure are encoded with invariant or similar attention patterns.
Empirically, we show that \modelname achieves stronger compositional generalization than the vanilla Transformer on multiple low-complexity semantic parsing and machine translation datasets.
In our analysis, we show that SoVQ indeed learns a syntactically clustered embedding space and SAL/SRL induces generalizable attention patterns, which lead to improved systematicity.\footnote{Our code is publicly available at \url{https://github.com/jiangycTarheel/SQ-Transformer}.}

\end{abstract}

\section{Introduction}

Natural languages demonstrate \textit{compositionality}, which states that the meaning of a complex expression is determined by its structure and the meanings of its lexical constituents~\cite{chomsky1957syntactic,montague1970universal}.
It leads to humans' algebraic capacity to \textit{systematically} understand a potentially infinite number of novel combinations of known structures and entities.
For example, someone who understands ``\textit{The cat is asleep}'' and ``\textit{The dog is awake}'' must simultaneously understand ``\textit{The dog is asleep}'' and ``\textit{The cat is awake}''.
Systematicity involves an intelligence's ability to process high-dimensional raw signals in a low-dimensional, abstract space and gives rise to its complex reasoning skills.

Early works argued that neural networks are associative devices that cannot capture compositionality~\cite{fodor1988connectionism,marcus1998rethinking} and are supported by the empirical results that a Transformer~\cite{vaswani2017attention} trained to parse ``\textit{walk twice}'', ``\textit{walk around left}'', and `\textit{jump}' fails to parse ``\textit{jump twice}'' and ``\textit{jump around left}'' in SCAN \addjump~\cite{lake2018generalization}.
Later works presented a more promising picture: for example, \citet{zhou-jiang-2023-datafactor} found that Transformers trained on an augmented, high-complexity dataset with more examples and diverse entities/structures
can systematically generalize to novel compositions in SCAN \addjump. 
Studies on large pretrained models~\cite{furrer2020compositional,drozdov2023compositional} also reveal their ability to systematically generalize.
However, data augmentation requires domain-specific knowledge and pretraining on large datasets is also prohibitively expensive.
Therefore, how to induce systematicity with low-complexity data has a significant value for improving the model's data efficiency, and remains an open and important research question.

\begin{figure*}[t]
\centering
\begin{minipage}{0.49\textwidth}
\centering
\includegraphics[width=0.75\linewidth]{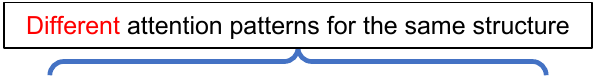}
\end{minipage}\hfill
\begin{minipage}{0.49\textwidth}
\centering
\includegraphics[width=0.7\linewidth]{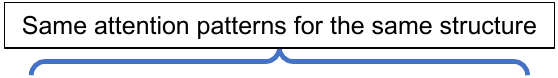}
\end{minipage}
\\
\vspace{-1em}
\subfloat[Attention maps encoding ``\textit{walk around left}'', trained on original \addjump.]{\label{subfig:1x_attn_maps_walkaroundleft} \includegraphics[width=0.23\textwidth]{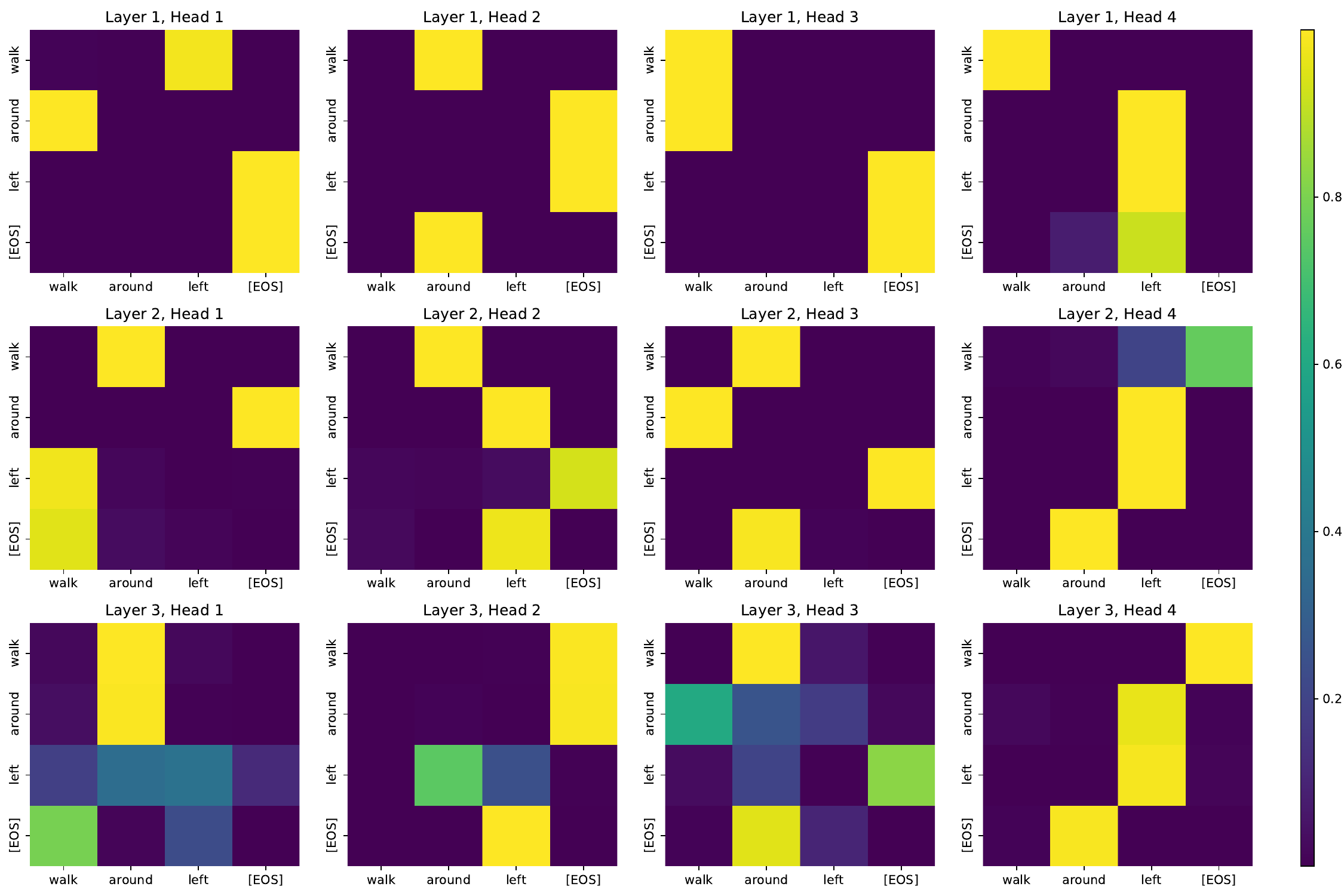}}
\hfill
\subfloat[Attention maps encoding ``\textit{jump around left}'', trained on original \addjump.]{\label{subfig:1x_attn_maps_jumparoundleft} \includegraphics[width=0.23\textwidth]{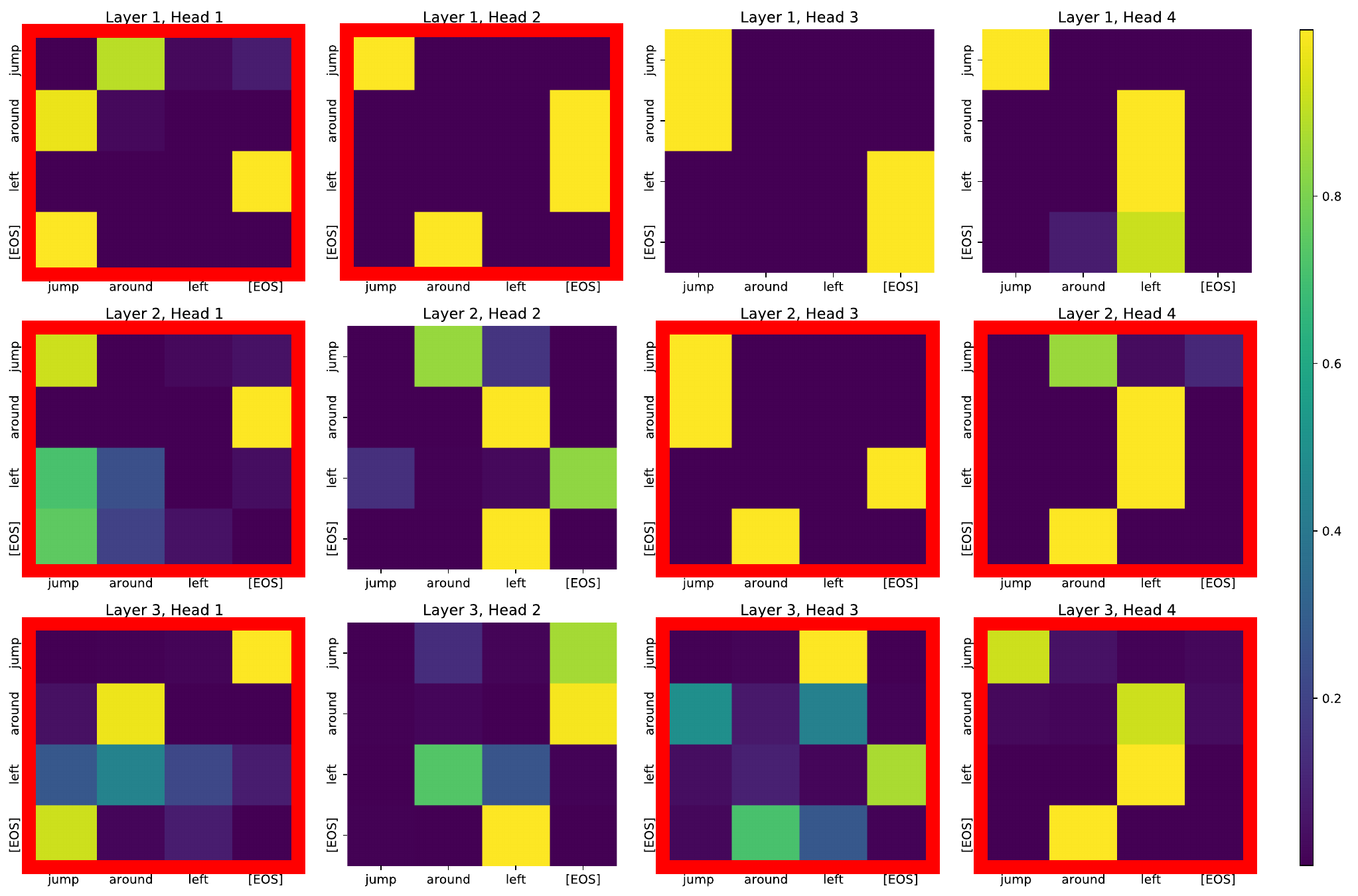}}%
\hfill
\subfloat[Attention maps encoding ``\textit{walk around left}'', trained on 20x augmented \addjump.]{\label{subfig:20x_attn_maps_walkaroundleft} \includegraphics[width=0.23\textwidth]{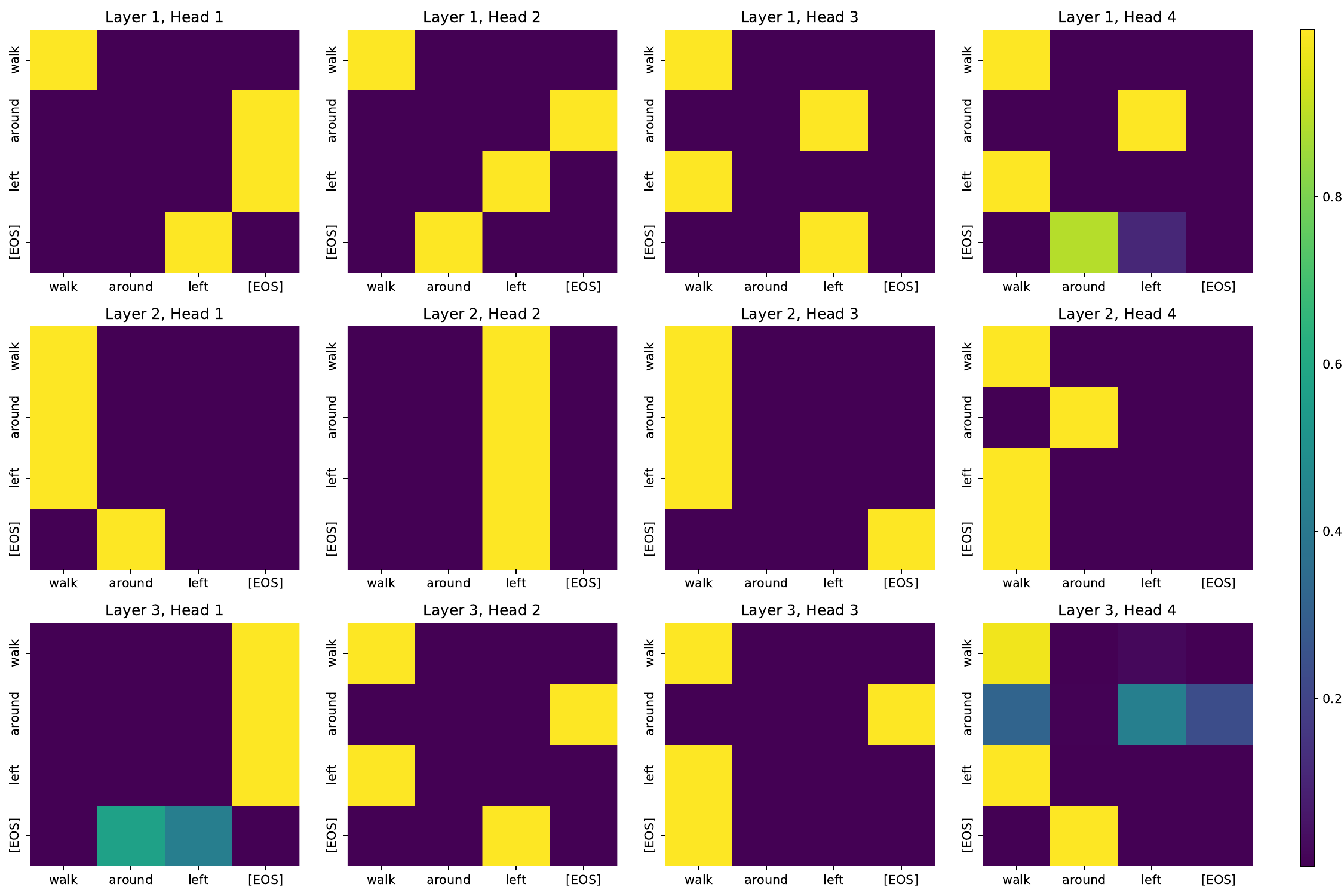}}
\hfill
\subfloat[Attention maps encoding ``\textit{jump around left}'', trained on 20x augmented \addjump.]{\label{subfig:20x_attn_maps_jumparoundleft} \includegraphics[width=0.23\textwidth]{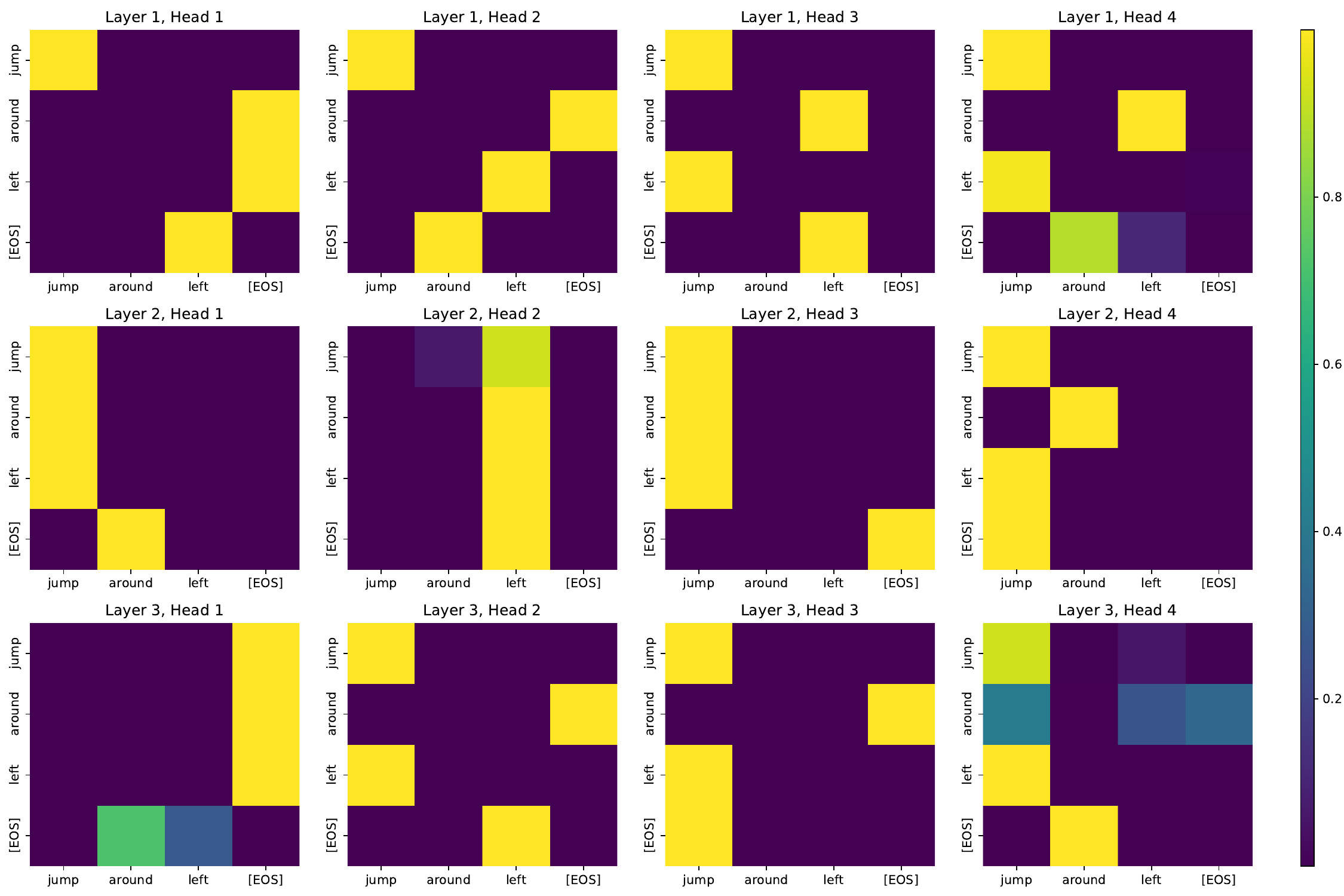}}%
\caption{Attention maps encoding a training example ``\textit{walk around left}'' and a test example ``\textit{jump around left}'' from the Transformers trained on the original SCAN \addjump training set (a and b), and 20x augmented training set (c and d) from~\citet{zhou-jiang-2023-datafactor} with 20 times more primitives like `\textit{walk1}' and more examples like ``\textit{walk1 around left}''.
We highlight the attention maps in (b) that differ from (a) in red boxes.
When trained on 20x augmented training set, the model encodes the two examples with highly similar attention maps across all layers and heads (c and d).
We show the attention maps of other training instances of the structure ``$\$x$ \textit{around left}'' in~\figref{fig:1x_20x_lookrunwalkjump_attn_maps}.
}
\label{fig:20x_attn_maps}
\end{figure*}

To understand the emergence of systematicity in Transformers, we start by analyzing the attention maps from models trained on SCAN \addjump data of different complexities. 
First, we demonstrate that a Transformer trained on the original, low-complexity training set (with only 4 primitives) uses different attention weights to encode in-distribution training sentences like ``\textit{walk around left}'' (\figref{subfig:1x_attn_maps_walkaroundleft}) and an unobserved sentence ``\textit{jump around left}'' (\figref{subfig:1x_attn_maps_jumparoundleft}).
It only achieves 3.7\% test accuracy in parsing unobserved sentences. 
On the other hand, the same model trained on large augmented data (with 84 distinct primitives like `\textit{walk}' and `\textit{jump}') uses highly similar attention patterns to encode these two sentences (as shown in~\figref{subfig:20x_attn_maps_walkaroundleft} and~\figref{subfig:20x_attn_maps_jumparoundleft}).
This model achieves 100\% test accuracy and encodes the structure ``$\$x$ \textit{around left}'' with a unified attention pattern that is invariant to the choice of $\$x$, as long as $\$x$ has the same syntactic function (e.g., being a verb).
Over the entire test set, we observe more of such reused attention from the model trained with augmented data.\footnote{We discuss a quantified analysis in~\secref{ssec:analysis_circuits}.}
Therefore, we test the hypothesis that 
this ability to systematically reuse learned attention patterns on novel sentences is critical for a Transformer to systematically generalize.

To this end, in this work, 
we propose \modelname with two improvements to the embeddings and attention layers respectively, 
so as to induce the same systematicity seen above, even with low-complexity training data.
The first improvement brings linguistic categorization to the word embeddings:
as \citet{johnson2004systematicity} argues \textit{``the claim that natural languages are systematic presupposes a natural non-overlapping linguistic categorization of all the expressions.''}
For example, for the model to generalize to the unseen ``\textit{jump twice}'', it first has to learn that `\textit{jump}' belongs to the same category as other primitives like `\textit{walk}' and `\textit{run}'.
Motivated by this theory, we propose \textbf{S}tructure-\textbf{o}riented \textbf{V}ector \textbf{Q}uantization (\textbf{SoVQ}) to actively cluster all word embeddings into a fixed number of structural equivalence classes, and quantize each word into a code embedding shared among an entire class.
We introduce a variational and generalized Brown Clustering~\cite{brown1992class} objective.
This unsupervised objective encourages a ``\textbf{predictive clustering}'' such that the class of a token can be predicted by the classes of its context tokens,
and hence ensures that words of the same syntactic function are in the same class.
After being trained on examples like ``\textit{walk}'', ``\textit{walk around left}'', and ``\textit{jump}'',
SoVQ can cluster `\textit{jump}' and `\textit{walk}' into the same class and quantize them into a shared code embedding that encodes their common structural information.

The second improvement encourages a unified attention pattern for encoding sentences of a common structure. 
The general belief in cognitive science states that \textit{systematicity involves a capacity to represent common structural relations among the equivalently cognizable entities}~\cite{phillips2016systematicity}.
That is, a systematic mind can always represent a structure even if one or more of its entities is substituted with any equivalently cognizable entity.\footnote{The notion of ``equivalently cognizable entities'' generally refers to entities within an equivalence class with respect to certain structural equivalence (e.g., all proper nouns).}
Since \textbf{SoVQ} has quantized each class of equivalently cognizable entities into a code embedding,
we then propose the \textbf{S}ystematic \textbf{A}ttention \textbf{L}ayer (\textbf{SAL}) that uses these code embeddings as the queries and keys, and the word embeddings as the values (\figref{fig:sal}).
When encoding sentences with a common structure like ``$\$x$ \textit{around left}'', SAL is \textbf{hard-invariant} for any $\$x$ in a structural equivalence class $C$ established by SoVQ.
It thus enables the Transformer to systematically represent common structural relations among those quantized classes of equivalently cognizable entities.

To retain the attention's ability to represent non-structural relations that commonly exist in natural languages, we also introduce an alternative to SAL: \textbf{S}ystematically \textbf{R}egularized \textbf{L}ayer (\textbf{SRL}).
It inherits the architecture of a regular attention layer, but additionally minimizes the L2 distance between the layers' outputs computed from word embeddings and the layers' outputs computed from quantized word embeddings (\figref{fig:srl}).
Therefore, unlike SAL, SRL encourages attention's \textbf{soft invariance} to structurally equivalent entities: sentences with common structures are processed with similar but not necessarily the same attention pattern.
Overall, we name the Transformer with the SoVQ and SAL/SRL as \textbf{\modelname} (Structurally Quantized Transformer). 

To demonstrate that predictive clustering in embeddings and invariance in attention can lead to systematicity in the model's predictions, 
we train and evaluate \modelname from scratch on multiple low-complexity semantic parsing and machine translation datasets requiring compositional generalization.
In semantic parsing, \modelname improves upon Transformer on SCAN \addjump x2~\cite{jiang-etal-2022-mutual} (40\%$\rightarrow$99.4\%), \aroundright~\cite{loula-etal-2018-rearranging} (69.5\%$\rightarrow$99.6\%), and COGS~\cite{kim-linzen-2020-cogs} (82.6\%$\rightarrow$83.4\%).
In machine translation, \modelname achieves higher BLEU scores (60.5$\rightarrow$62.8) and lower novel compound translation error (29.6\%$\rightarrow$18.1\%) on CoGnition~\cite{li-etal-2021-compositional}.
Interestingly, it also shows generalizability to higher-complexity, natural datasets that do not have a significant distribution shift between training and test sets:
in WMT En$\leftrightarrow$De and En$\leftrightarrow$Fr, \modelname with SRL obtains significantly higher BLEU scores.
We further analyze \modelname and present two findings:
(1) SoVQ can more effectively cluster word embeddings based on their syntactic functions compared to VQ;
(2) SAL and SRL learn attention patterns that can systematically encode unseen compositions of structure and entities.
These analyses explain the working mechanism of \modelname and verify our insights in designing these modules.

In summary, we propose \modelname, which quantizes word embeddings based on their syntactic functions and learns generalizable attention for sentences of the same structure.
We hope this work can shed more light on the inner mechanism of Transformers' generalization and lay the groundwork for future improvement in architecture design and representation learning.

\section{Background and Motivation}
\label{sec:bg}

\paragraph{Vector Quantization}
(VQ)~\cite{agustsson2017soft,van2017vqvae} is a compression technique that represents a set of representations $e_x$ of the variable $x$ by a small, fixed number of code embeddings $\mathbf{z}$.
The code is inferred with the nearest neighbor look-up on a codebook $Z\in R^{K\times D}$ made up of $K$ embeddings of the dimension $D$:
\begin{equation}
\label{eqn:quantization}
\begin{split}
q(z_k|x) &= 
\begin{cases}
1 \; \text{if} \; k = \mathop{\mathrm{argmin}}_{j} \;\mathrm{f}(e_x, z_j) \\
0 \; \text{otherwise} \\
\end{cases} \\
\mathrm{VQ}(x) &= z_k \;\text{where}\; q(z_k|x) = 1 \\
\end{split}
\end{equation}
where $\mathrm{f}$ is a distance function (e.g., negative cosine similarity).
The discrete code embeddings are updated using exponential moving averages of $e_x$.
Previous works~\cite{van2017vqvae,razavi2019vqvae2,ramesh2021dalle} have shown that VQ-VAE can generate high-fidelity, continuous signals like images and speech.
The exploration of vector quantization on modeling languages~\cite{lingle2023transformer} remains limited, primarily because (sub)words are already discrete features.
In this work, we use VQ to cluster words based on their syntactic function (\secref{ssec:method_vq}). 

\paragraph{Brown Clustering}

\cite{brown1992class} is a word clustering algorithm that divides a vocabulary $V$ into $m$ mutually exclusive classes by maximizing the mutual information $I(Z_1,Z_2)$ between the classes of a random bigram $(X_1,X_2)$ in a sentence:
\begin{equation}
\label{eqn:brown_lustering_1}
\begin{split}
\max_{Z:V\rightarrow[m]} &= \sum_{z_1,z_2} \frac{\#(z_1, z_2)}{N}\mathrm{log}(\frac{\#(z_1,z_2)N}{\#(z_1)\#(z_2)})\\
\end{split}
\end{equation}
where $\#(z, z')$ denotes the number of occurrences of the cluster pair $(z, z')$ for any bigram in $[x_1 . . . x_N ]$.\footnote{By assuming a uniform distribution over consecutive word pairs $(x_{i-1}, x_i)$, \citet{brown1992class} approximate $p(z_1,z_2)$ and $p(z)$ using $\frac{\#(z_1,z_2)}{N}$ and $\frac{\#(z)}{N}$ to derive~\eqnref{eqn:brown_lustering_1}.}
This algorithm can cluster a vocabulary based on the syntactic functions of words by promoting ``predictive clustering'': the class of a token must be predictable from the class of its context token.
However, it requires nontrivial combinatorial optimization and is difficult to scale and generalize for modern neural networks.
In this work, we propose a variational objective of Brown Clustering (\secref{sssec:variational_brown}) that can be optimized with gradient descent.
\section{\modelname}

In this section, we introduce the components of \modelname: 
(1) Structural-oriented Vector Quantization (\secref{ssec:method_vq});
(2) Systematic Attention Layer that operates on quantized embeddings (\secref{ssec:method_sal});
and (3) Systematically Regularized Layer that regularizes the attention outputs (\secref{ssec:method_srl}).

\paragraph{Notations.}

We denote the source and target sequences as $[x_i]$ and $[y_j]$.
The seq2seq framework consists of an encoder with word embeddings $E_x$ and a decoder with word embeddings $E_y$.
For quantizing $E_x$ and $E_y$, we define two codebooks $Z_x$ and $Z_{y}$ with $K_x$ and $K_y$ code embeddings respectively.

\subsection{Structure-oriented Vector Quantization}
\label{ssec:method_vq}
Same as the original VQ, Structure-oriented Vector Quantization (SoVQ) clusters the (sub)word embeddings into several classes and quantizes embeddings within a class to a shared code embedding (\eqnref{eqn:quantization}).
We discuss a previous MMI objective and then propose variational Brown Clustering that better cluster words based on their syntactic functions.

\subsubsection{Variational MMI objective}
\citet{stratos-2019-mutual} proposed an unsupervised part-of-speech tagging method by maximizing the mutual information (MMI) between the inferred class $Z$ of a token $X$ and its surrounding context $\hat{X}$.
It defines $q(z|x)$ that directly infers the class of $x$ (posterior) and $p(z|\hat{x})$ that predicts the cluster of $x$ based on its context $\hat{x}$ (prior).
It maximizes the variational lower bound of the mutual information $I(\hat{X},Z)$:
\begin{equation}
\label{eqn:elbo_mmi}
\begin{split}
I(\hat{X},Z) &= H(Z) - H(Z|\hat{X}) \\
&\geq H(Z) - H(q,p)\\
\end{split}
\end{equation}
where $H(q,p)$ is the cross entropy over samples:
\begin{equation}
\label{eqn:h_q_p}
\begin{split}
H(q,p) &= \mathop{\mathbb{E}}_{x,\hat{x}\sim D}\left[-\sum_{z}q(z|x)\mathrm{log}p(z|\hat{x})\right] \\
\end{split}
\end{equation}
We show more details the derivation of this lower bound in~\appendixref{appendix_method:elbo_deduction}.
As we can see in~\eqnref{eqn:elbo_mmi}, maximizing this ELBo is equivalent to (1) minimizing cross-entropy between the cluster inference posterior $q(z|x)$ and cluster prediction prior $p(z|\hat{x})$ and (2) maximizing the entropy $H(Z)$ of the cluster distribution. 

\textit{First, minimizing the cross-entropy $H(q,p)$ enforces ``predictive clustering'': the class of $x$ must be predictable from its context $\hat{x}$.}
We introduce a theorem to show how this leads to assigning words appearing in the same context to the same class.

\begin{theorem}
\label{theo:cross_entropy}
    Let $x_a$ and $x_b$ be two tokens that only appear in the same sets of context $\hat{X}$. 
    Let $p',q'=$ \[\mathop{\arg\min}_{p,q} H(q(z|x_a), p(z|\hat{x})) + H(q(z|x_b), p(z|\hat{x}))\] Then, we have:
    $q'(z|x_a) = q'(z|x_b)\;\;\forall z\in Z$, which means $x_a$ and $x_b$ are clustered into the same class in the optimal solution.
\end{theorem}
The proof is straightforward: the minimum value of the cross-entropy loss is 0, and this can only be achieved when $q(z|x_a) = q(z|x_b) = p(z|\hat{x})$.
To better understand the mathematical intuition of this theorem, consider the case where all adjectives are categorized into the same cluster.
Then, we can confidently predict the class of $\$x$=`\textit{amazing}' based on the context ``\textit{The food tastes} $\$x$.'', thus achieving the low cross-entropy with the posterior.

\textit{Second, maximizing the entropy of the cluster distribution pushes the model to utilize every cluster $z$ in the latent space with (almost) equal probability.}
It thus prevents the trivial solution that assigns all tokens to only one random cluster $k$: $p(z_k|x)=q(z_k|x)=1$
to minimize the first cross-entropy term ($H(p,q)=0$).
Empirically, this variational MMI objective achieves strong unsupervised POS tagging performance~\cite{stratos-2019-mutual}.

\begin{figure*}[t]
\centering
% \hspace*{\fill}
\hspace{-20pt}
\subfloat[The Systematic Attention Layer (SAL).]{\label{fig:sal} \includegraphics[width=0.4\textwidth]{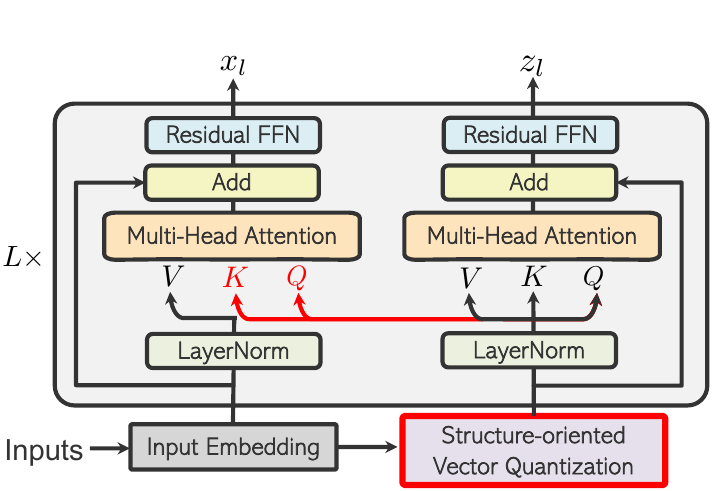}}
% \hfill
\hspace{40pt}
\subfloat[The Systematically Regularized Layer (SRL).]{\label{fig:srl} \includegraphics[width=0.4\textwidth]{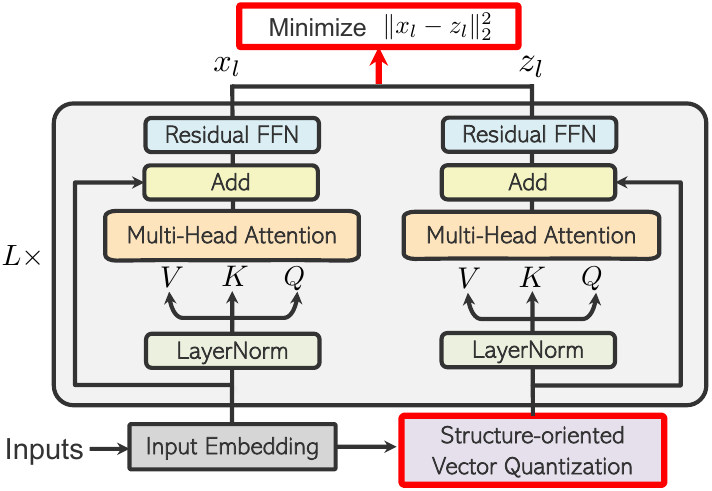}}
  % \hspace*{\fill}
\caption{Architecture of the Systematic Attention Layer (SAL) and the Systematically Regularized Layer (SRL). 
}
\label{fig:sal_srl}
\end{figure*}

\subsubsection{Variational Brown Clustering}
\label{sssec:variational_brown}
In this work, we propose another MMI objective that marries the original Brown Clustering objective $I(Z_1, Z_2)$ and the variational MMI (\eqnref{eqn:elbo_mmi}).
First, we redefine the cluster prediction distribution as $p(z|\hat{z})$, where $\hat{z}$ are the quantized codes of all context tokens $\hat{x}$ inferred from $\mathrm{argmax}(q(z|\hat{x}))$.
This differs from the $p(z|\hat{x})$ that predicts the cluster of $x$ directly from its context $\hat{x}$.
Then, instead of maximizing the ELBO of $I(\hat{X}, Z)$, we maximize the ELBO of $I(\hat{Z}, Z)$:
\begin{equation}
\label{eqn:elbo_mmi_reimagined}
\begin{split}
I(\hat{Z},Z) &= H(Z) - H(Z|\hat{Z}) \\
&\geq H(Z) - H(q(z|x),p(z|\hat{z}))\\
\end{split}
\end{equation}
This inequality is still valid\footnote{We show the derivation in~\appendixref{appendix_method:reimagined_brown_clustering}.} even though we replaced $\hat{X}$ and $p(z|\hat{x})$ in~\eqnref{eqn:elbo_mmi} with $\hat{Z}$ and $p(z|\hat{z})$.
The objective $I(\hat{Z}, Z)$ becomes the exact Brown Clustering objective if we set $\hat{x}$ as a random context token rather than all of them.
We show the implementation of this variational loss in~\appendixref{appendixssec:implementation}.

We argue that this variational Brown Clustering objective can better cluster words based on their syntactic functions than the lower bound of $I(\hat{X},Z)$ (\eqnref{eqn:elbo_mmi}).
This is because, according to~\theoremref{theo:cross_entropy}, maximizing $I(\hat{X},Z)$ can only cluster words that appear in similar contexts into the same class.
However, some words having the same syntactic function might rarely occur in the same context due to semantics.
For example, `\textit{police}' and `\textit{professor}' usually appear in very different contexts: ``\textit{The police arrested a thief.}'' and ``\textit{The professor appraised a student.}''
Therefore, maximizing $I(\hat{X},Z)$ might not push the model to assign them to the same cluster.
In comparison, maximizing $I(\hat{Z},Z)$ (Brown Clustering objective) can encourage clustering structurally equivalent words that appear in various contexts together: even though `\textit{police}' and `\textit{professor}' have different $\hat{X}$, they share the same $\hat{Z}$ given a structure-oriented word cluster.\footnote{The necessary clustering scheme that can achieve the purpose is [\{\textit{arrested,appraised}\}, \{\textit{thief,student}\}]}.
We support this claim with ablations in~\secref{table:ablation}.

\subsection{Systematic Attention Layer}
\label{ssec:method_sal}
Now that we have quantized each word embedding into a code $z$ encoding its syntactic functions,
we then use the quantizations as the queries and keys in computing the attention weights in the first layer.
Here we show the encoder's self-attention module (visualized in~\figref{fig:sal}):
\begin{equation*}
\begin{split}
z_{l+1} &= \mathrm{MHAttn}(q=z_l,k=z_l,v=z_l) \\
x_{l+1} &= \mathrm{MHAttn}(q=z_l,k=z_l,v=x_l) \\
\end{split}
\end{equation*}
where $x_0$ and $z_0$ are the non-contextualized word embeddings and their quantized code embeddings respectively.
The two attention modules ($\mathrm{MHAttn}$) share all parameters.
We call it the Systematic Attention Layer (SAL)
because this modified attention module promotes the systematic reusing of attention patterns:
as words of the same syntactic function (e.g., `\textit{cat}' and `\textit{dog}', `\textit{asleep}' and `\textit{awake}') are in the same cluster and hence share the same code embedding, the Transformer would process two sentences of the same syntactic structure (``\textit{The cat is asleep}'' and ``\textit{The dog is awake}'') using the same attention weights at every head and layer.
As a result, a model that understands one sentence is more likely to generalize to the other one, which is the key ability stemming from a systematic language understanding.
Similarly, we also use SAL with a regular cross attention in the decoder, please see~\appendixref{appendix_ssec:sal_decoder} for details.
In summary, SAL enforces \textbf{hard attention invariance} among sentences of the same syntactic structure, but at the cost of the flexibility of encoding non-structural relations that commonly exist in natural languages (e.g., idioms, commonsense, etc).
We discuss these cases in~\secref{ssec:discussion_compositionality}.

\subsection{Systematically Regularized Layer}
\label{ssec:method_srl}

To encourage systematicity in attention while keeping its ability to encode non-structural relations, 
we instead use the attention outputs computed from the quantized embeddings to regularize the attention outputs computed from word embeddings, 
by minimizing their squared L2 distances (MSE):
\begin{equation*}
\begin{split}
z_{l+1} &= \mathrm{MHAttn}(q=z_{l},k=z_{l},v=z_{l}) \\
x_{l+1} &= \mathrm{MHAttn}(q=x_{l},k=x_{l},v=x_{l}) \\
\mathcal{L} &\mathrel{+}= \beta\cdot\sum_{l=1}^{L}\lVert x_{l} - z_{l} \rVert^2_2 \\
\end{split}
\end{equation*}
where $x_0$ are the word embeddings. 
We name it Systematically Regularized Layer (SRL) and visualize it in~\figref{fig:srl}.
Unlike SAL, SRL demonstrates ``\textbf{soft invariance}'' so that sentences of a common structure are processed with similar but not necessarily the same attention pattern.
\section{Experiments}

\subsection{Datasets}
We use a series of semantic parsing and machine translation tasks requiring compositional generalization and the common WMT tasks.

\paragraph{SCAN \addjump}
\cite{lake2018generalization} tests the models' ability to generalize syntactic structures (e.g., ``$\$x$ twice'') to a novel entity ($\$x$ = `\textit{jump}').
Here we use the augmented training set with 2 times more primitives~\cite{jiang-etal-2022-mutual}.

\paragraph{SCAN \aroundright}
\cite{loula-etal-2018-rearranging} tests the models' ability to generalize a common syntactic structure ``$\$x_1$ around $\$x_2$'' to an entity ($\$x_2$=`\textit{right}') that is only associated with other structures during the training.

\paragraph{COGS}
\cite{kim-linzen-2020-cogs} challenges models to parse a diverse set of natural language sentences into their corresponding logical forms based on lambda calculus to accurately reflect the semantic representation of the natural sentence. 

\paragraph{CoGnition}
\cite{li-etal-2021-compositional} is an English-to-Chinese translation dataset with a synthetic OOD test set, where each sentence contains novel compositions of common structures and constituents from the training set.

\paragraph{WMT.}
We use WMT17 English$\leftrightarrow$German~\cite{wmt17} and WMT14 English$\leftrightarrow$French~\cite{wmt14} translation tasks.

\begin{table}[t!]
\centering
\begin{small}
\begin{tabular}[t]{c|ccc}
\toprule
 \centering \textbf{Model} & \textsc{Jump} 2x & \textsc{AroundR} & \textsc{COGS}\\
\midrule
\multicolumn{4}{c}{\multirow{1}{*}{\textsc{Previous Models}}} \\
\midrule
 LSTM-RNN & 1.2 & 2.5\tiny{$\pm$2.7} & - \\
 CGPS-RNN & 98.8\tiny{$\pm$1.4} & 83.2\tiny{$\pm$13.2} & - \\
 Lex Learn & 92.0\tiny{$\pm$0.2} & 95.0\tiny{$\pm$0} & 82.0\tiny{$\pm$0} \\
\midrule
\multicolumn{4}{c}{\multirow{1}{*}{\textsc{Our Models}}} \\
\midrule
 Transformer & 40.04\tiny{$\pm$17.3} & 69.47\tiny{$\pm$9.2} & 82.60\tiny{$\pm$0.5}\\
 \modelname  & 99.42\tiny{$\pm$1.0}$^{\star}$ & \textbf{99.63}\tiny{$\pm$0.6}$^{\star}$ & \textbf{83.36}\tiny{$\pm$0.7} \\
\bottomrule
\end{tabular}
\caption{Test accuracy from the SCAN \textsc{AddJump} (2x augmented), \textsc{AroundRight}, and COGS.
We report previous models: LSTM-RNN~\cite{lake2018generalization}, CGPS-RNN~\cite{li-etal-2019-compositional}, and Lex Learn~\cite{akyurek-andreas-2021-lexicon}.
We report our models' average ($\pm$ std.) results from 5 random seeds. 
\modelname results with $^{\star}$ use SAL while others use SRL.
}
\label{table:scan_results}
\end{small}
\end{table}

\subsection{Results}
\paragraph{Semantic Parsing results.}
We show the experimental setup in~\appendixref{appendix_ssec:experimental_setup}.
We report the results on the two SCAN tasks and COGS in~\tabref{table:scan_results}.
Specifically, \modelname with SAL achieves significant\footnote{Bootstrapped test with $\alpha<0.01$.} improvements over the baseline on SCAN \addjump and \aroundright.
With SRL, \modelname manages to outperform the baseline on the larger, more natural COGS dataset.
This shows the effectiveness of \modelname in generalizing to unseen combinations of syntactic structure and lexical constituents.
We compare the performance of SAL and SRL on a small, synthetic dataset and a larger, natural dataset in~\secref{ssec:ablation}.

\begin{table}[t!]
\centering
\begin{small}
\begin{tabular}[t]{c|cc|c}
\toprule
 \multicolumn{1}{c}{\multirow{2}{*}{\textbf{Model}}} & \multicolumn{2}{c}{\multirow{1}{*}{CTER ($\downarrow$)}} & \multicolumn{1}{c}{\multirow{2}{*}{BLEU}} \\
 \cmidrule{2-3}
  & Instance & Aggregate & \\
\midrule
\multicolumn{4}{c}{\multirow{1}{*}{\textsc{Previous Models}}} \\
\midrule
 Transformer & 28.4 & 62.9 & 59.5 \\
 Proto-Transformer & 21.7 & 51.8 & 60.1 \\
 Dangle-Transformer & 22.8 & 50.6 & 60.6 \\
 Consistency-Reg & \underline{20.2} & \textbf{48.3} & \underline{61.3} \\
\midrule
  \multicolumn{4}{c}{\multirow{1}{*}{\textsc{Our Models}}} \\
\midrule
 Transformer & 29.55 & 61.62 & 60.45  \\
 \modelnamenospace\tiny{$_{\text{SRL}}$} & \textbf{18.14} & \underline{48.89} & \textbf{62.78} \\
\bottomrule
\end{tabular}
\caption{Compound Translation Error Rate (CTER, lower is better) and BLEU score on the Compositional Generalization test set from the CoGnition En-Zh.
We also report the results from Proto-Transformer~\cite{yin-etal-2022-categorizing}, Dangle-Transformer~\cite{zheng-lapata-2022-disentangled}, and consistency-regularized Transformer~\cite{yin-etal-2023-consistency}.
The best result is \textbf{bold} and the 2nd best is \underline{underlined}.
}

\label{table:cognition}
\end{small}
\end{table}

\begin{table}[t!]
\centering
\begin{small}
\begin{tabular}[t]{c|cccc}
\toprule
\textbf{Model} & En-De & De-En & En-Fr & Fr-En \\
\midrule
 Transformer & 28.10 & 31.30 & 37.01 & 34.24 \\
 \modelnamenospace\tiny{$_{\text{SRL}}$} & \textbf{29.21} & \textbf{31.96} & \textbf{38.38} & \textbf{35.56} \\
\bottomrule
\end{tabular}
\caption{BLEU scores on WMT17 En$\leftrightarrow$De and WMT14 En$\leftrightarrow$Fr test sets.
}
\label{table:wmt}
\end{small}
\end{table}

\paragraph{Machine Translation results.}
We evaluate the baseline Transformer as well as \modelname on the CoGnition compositional generalization test set and WMT test sets and report their BLEU 4~\cite{papineni2002bleu} scores.
For CoGnition, we also report the novel compound translation error (CTER)~\cite{li-etal-2021-compositional}.
It examines whether all of the atoms (tokens) in the novel compound are correctly translated in the generated Chinese sentence.
Specifically, instance-level CTER denotes the percentage of the test instances in which one or more atoms in the novel compound are translated incorrectly.
Aggregate-level CTER denotes the percentage of novel compounds that are translated wrong in at least one instance.

Compared to the Transformer baseline, \modelname obtains significantly higher BLEU scores on CoGnition En$\rightarrow$Zh (\tabref{table:cognition}), WMT17 En$\leftrightarrow$De, and WMT14 En$\leftrightarrow$Fr tasks (\tabref{table:wmt}).
On CoGnition, \modelname achieves substantially lower instance and aggregate compound error rate in its Chinese translation.
This improvement shows that SoVQ and SRL enable the model to correctly translate more novel compounds.

\begin{figure}[t]
\centering
\subfloat[Source embeddings trained with no quantization.]{\label{subfig:scan_src_baseline} \includegraphics[width=0.23\textwidth]{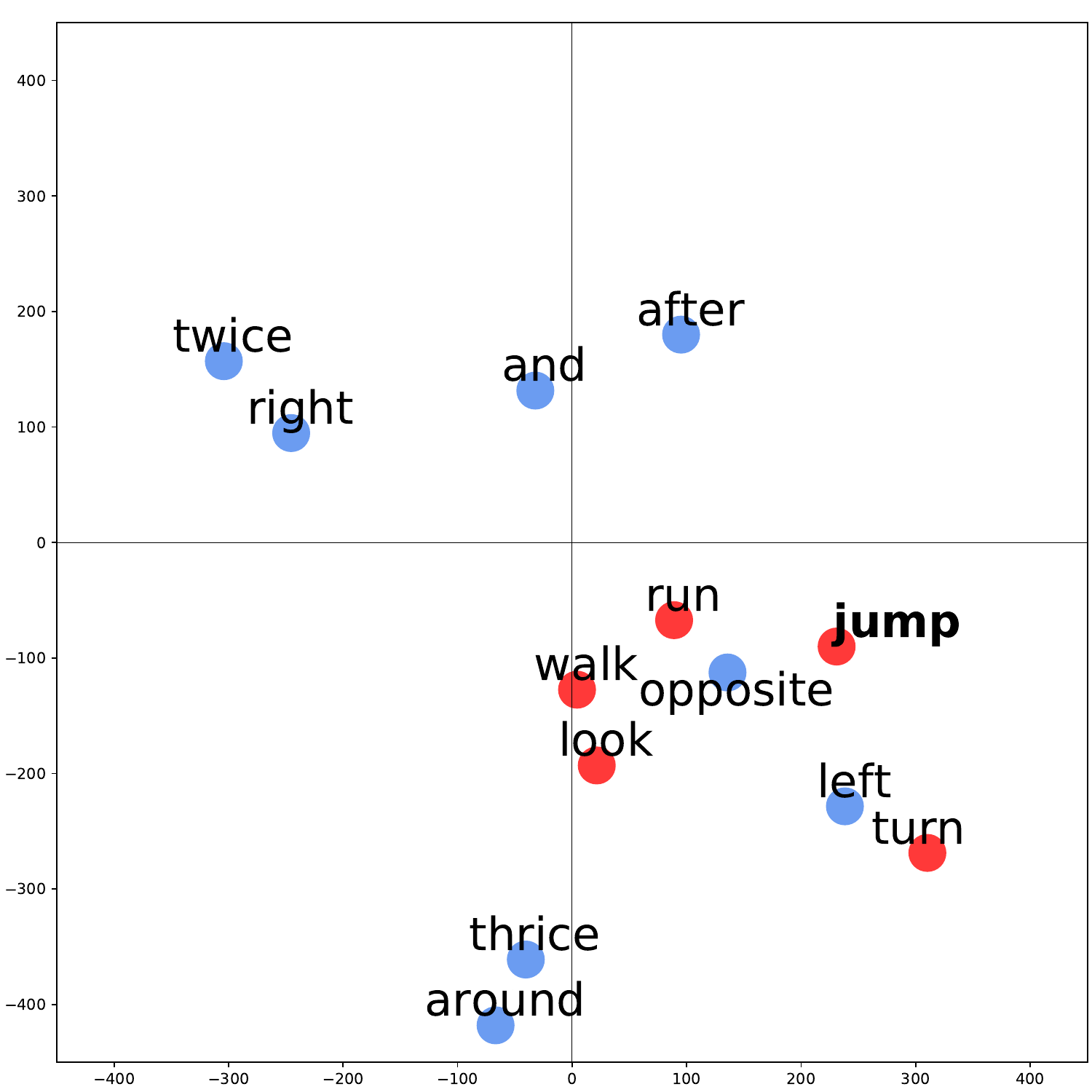}}
\hfill
\subfloat[Source embeddings trained with SoVQ (6 classes).]{\label{subfig:scan_src_quantized} \includegraphics[width=0.23\textwidth]{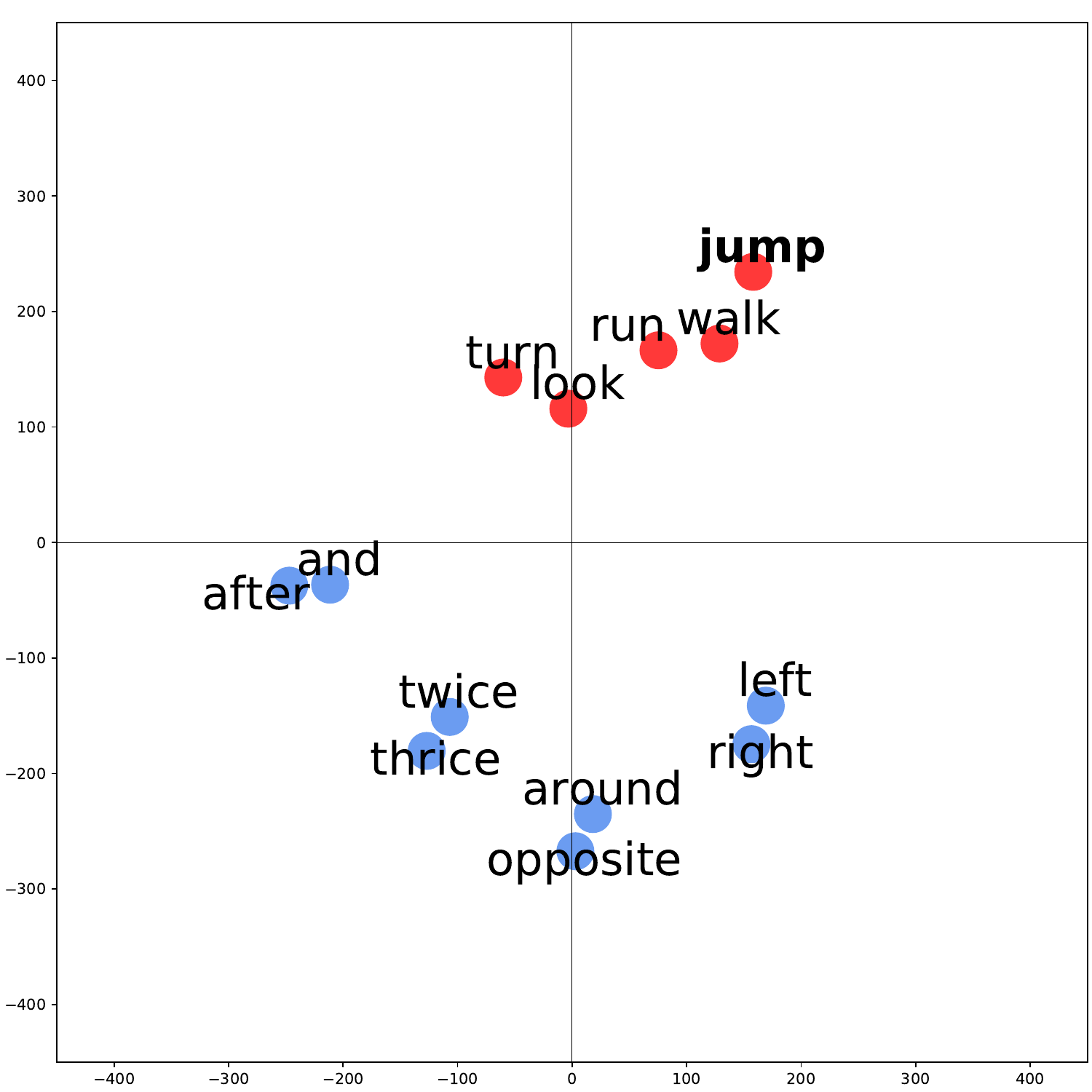}}%
\hfill
\caption{T-SNE visualization of embeddings learned on \textsc{SCAN AddJump} dataset~\cite{lake2018generalization}. 
}
\label{fig:tsne_scanjump}
\end{figure}

\begin{figure*}[t]
\centering
\subfloat[$n_a$=1 (acc=2.08)]{\label{subfig:scan_src_jump1} \includegraphics[width=0.23\textwidth]{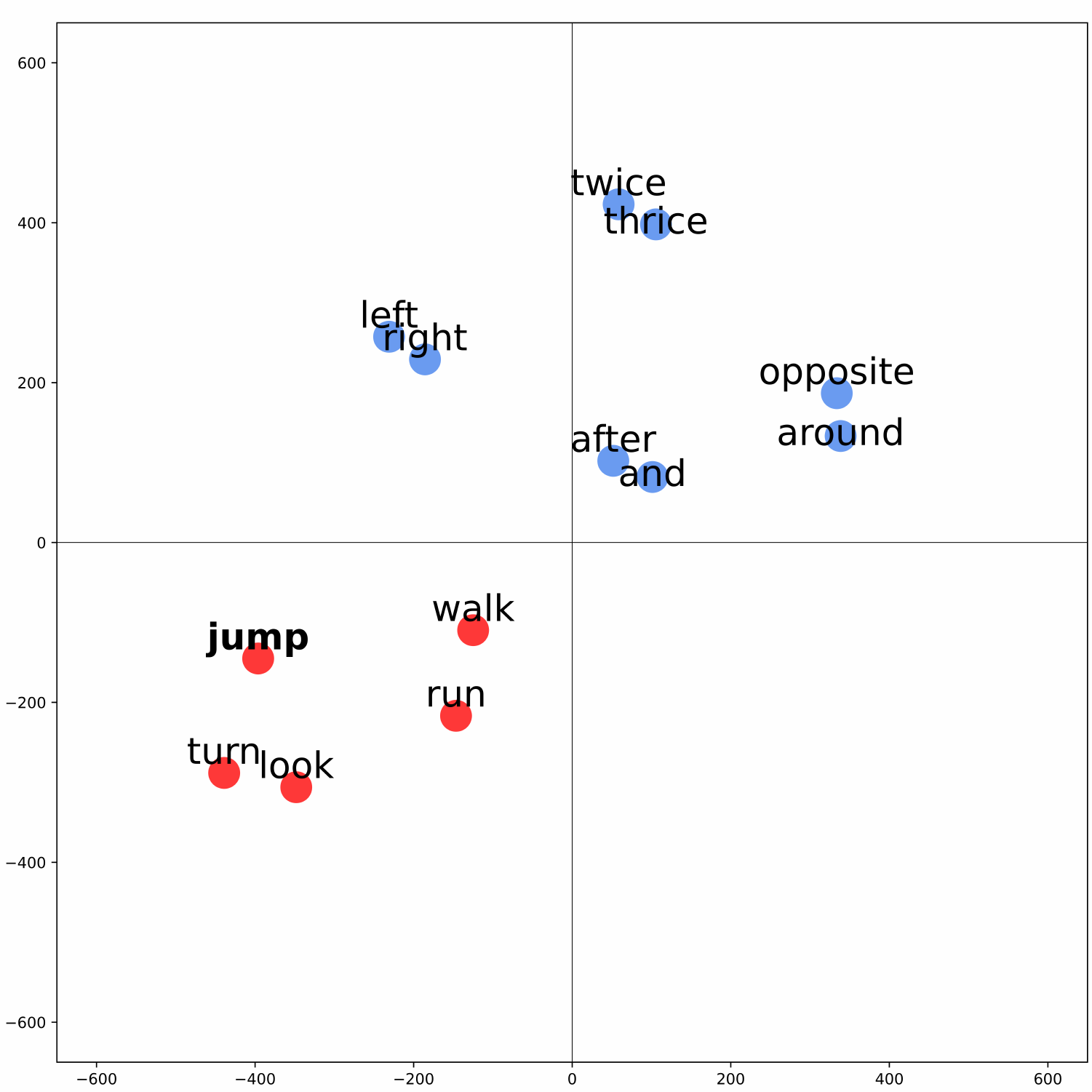}}
\hfill
\subfloat[$n_a$=10 (acc=4.68)]{\label{subfig:scan_src_jump10} \includegraphics[width=0.23\textwidth]{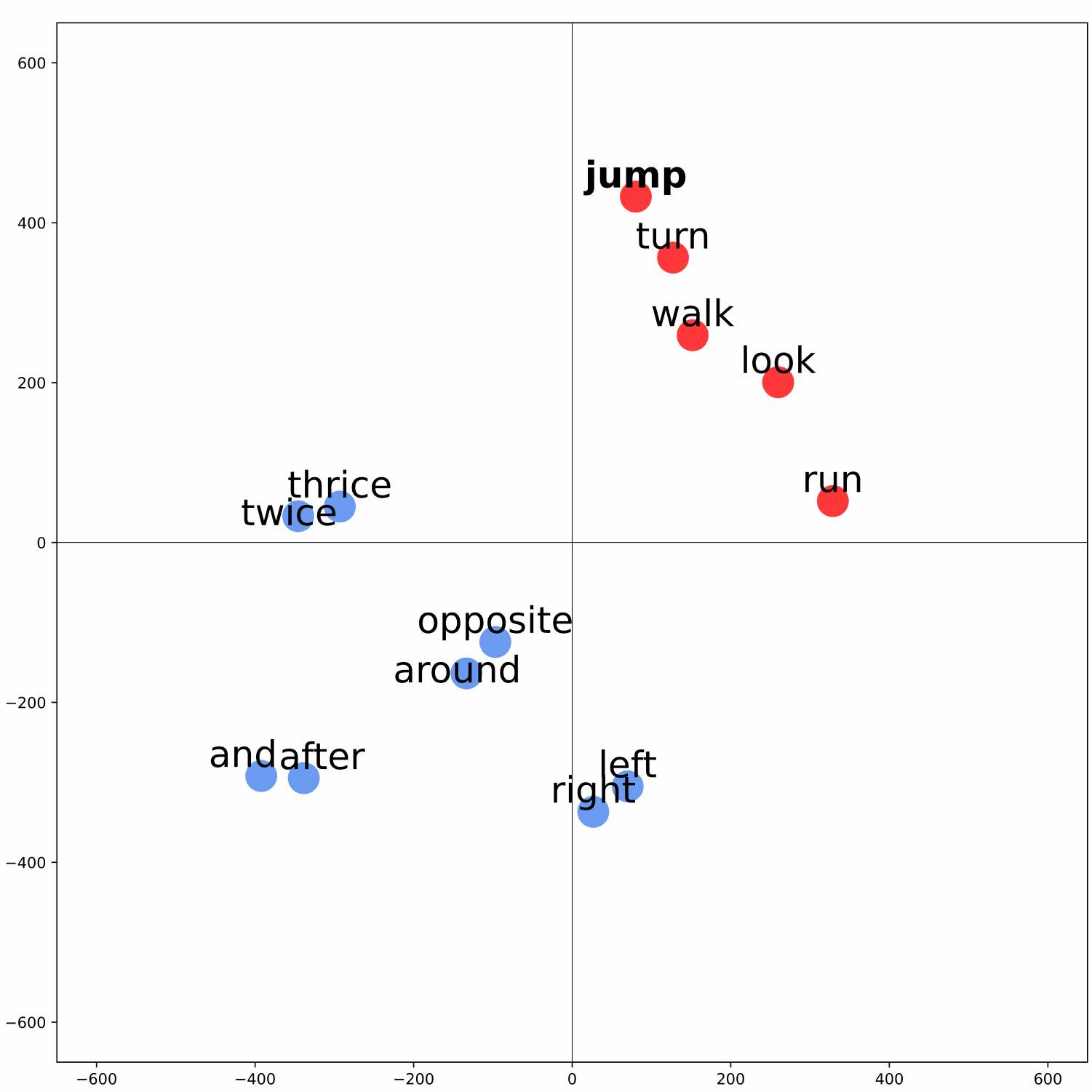}}%
\hfill
\subfloat[$n_a$=100 (acc=79.38)]{\label{subfig:scan_src_jump100} \includegraphics[width=0.23\textwidth]{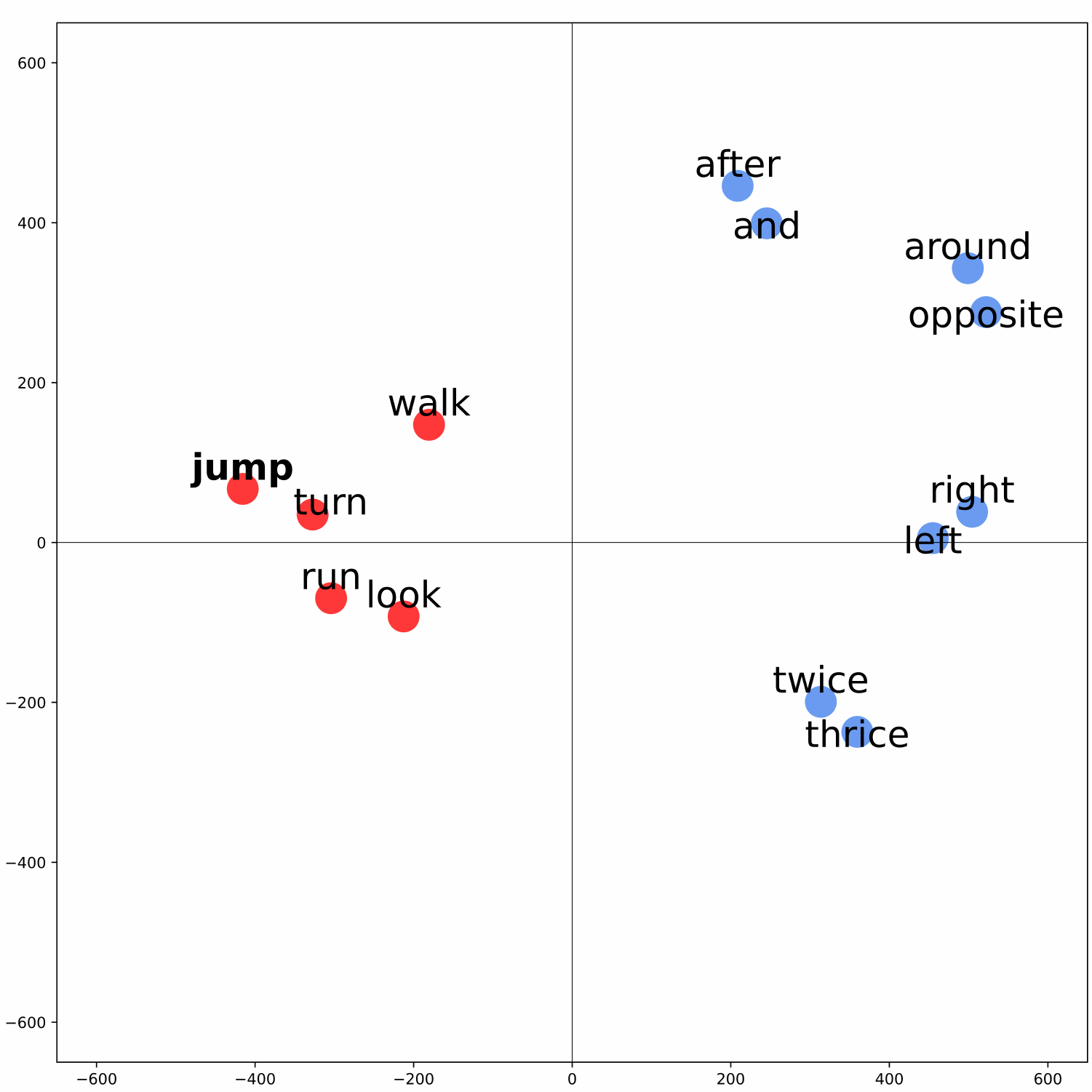}}%
\hfill
\subfloat[$n_a$=1000 (acc=100.00)]{\label{subfig:scan_src_jump1000} \includegraphics[width=0.23\textwidth]{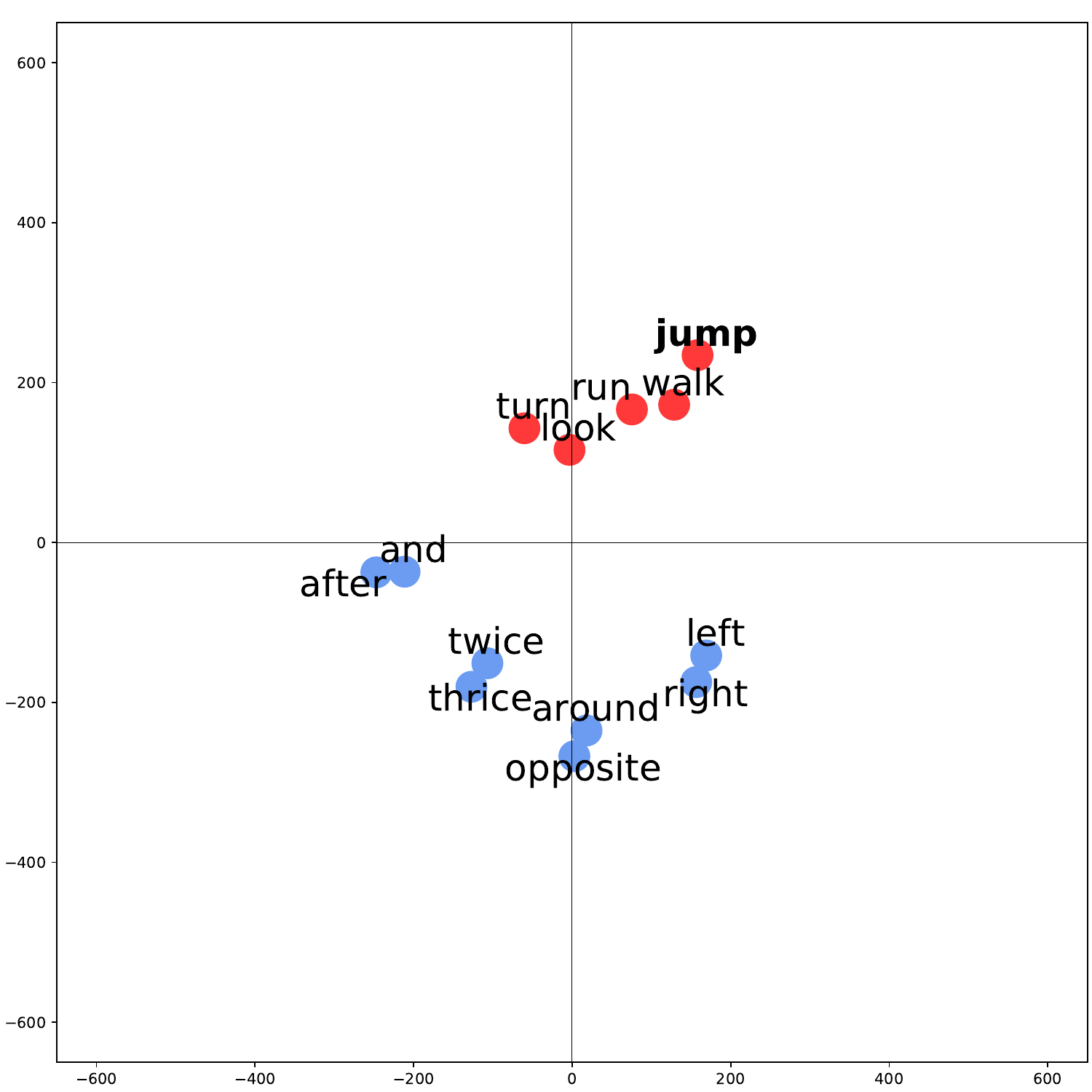}}%
\caption{T-SNE visualization of embeddings learned on \textsc{SCAN AddJump} dataset~\cite{lake2018generalization} with $n_a$ atomic expressions (e.g., \textit{jump}$\mapsto$\texttt{JUMP} and \textit{walk}$\mapsto$\texttt{WALK}) for each primitive and the model's accuracy (acc).}
\label{fig:ablation_num_atomic}
\end{figure*}

\begin{table}[t!]
\centering
\begin{small}
\begin{tabular}[t]{l|cc}
\toprule
 \centering \modelname & \addjump 2x & CoGnition\\
\midrule
 w. SAL & \textbf{99.42}\tiny{$\pm$0.98}$^{\star}$ & 59.85\tiny{$\pm$0.49}$^{\star}$\\
 w. SRL & 47.36\tiny{$\pm$20.83}$^{\dagger}$ & \textbf{62.35}\tiny{$\pm$0.52}$^{\dagger}$ \\
  None & 53.79\tiny{$\pm$18.36} & 61.11\tiny{$\pm$0.34} \\
\midrule
 - SoVQ & 78.44\tiny{$\pm$34.01}$^{\star}$ & 60.93\tiny{$\pm$0.13}$^{\dagger}$ \\
 - Brown & 97.75\tiny{$\pm$3.93}$^{\star}$ & 61.52\tiny{$\pm$0.22}$^{\dagger}$ \\ 
\bottomrule
\end{tabular}
\caption{Ablation: test accuracy on SCAN \addjump (2x augmented) and BLEU on CoGnition (averaged over 5 runs). 
`- SoVQ' uses the original Vector Quantization.
`- Brown' uses the MMI objective from~\citet{stratos-2019-mutual} instead of our variational Brown Clustering objective.
Results with $^{\star}$ use SAL while results with $^{\dagger}$ use SRL.
}
\label{table:ablation}
\end{small}
\end{table}

\section{Analysis}
In this section, we conduct an ablation study (\secref{ssec:ablation}) and analyze the embeddings (\secref{ssec:analysis_embedding}) as well as the attention patterns (\secref{ssec:analysis_circuits}).

\subsection{Ablation Study}
\label{ssec:ablation}
We conduct an ablation study on the components of \modelname and present the results in~\tabref{table:ablation}.
Most notably, we find that SAL only works on small, synthetic datasets like SCAN \addjump but not on CoGnition;
while SRL only works on larger, more natural datasets like CoGnition and COGS but not on SCAN tasks.
This finding is in line with other works~\cite{li-etal-2019-compositional,russin2020systematicity} that only achieve strong performance on small, synthetic datasets.
It suggests that generalizing on SCAN requires learning its grammar from an extreme data setting, which might only be possible via some strictly structural inductive biases like hard invariance to primitive substitution.
Generalizing on natural languages, on the other hand, requires learning both structural and some non-structural relations (discussed in~\secref{ssec:discussion_compositionality}), which is not possible using SAL that is hard invariant under a fixed structure.
Later (in~\secref{ssec:analysis_circuits}), we will show how SRL injects a similar, but soft invariance into attention to perform strongly on more natural tasks.

Moreover, we show that removing either SAL/SRL (`None') or SoVQ (`-SoVQ') results in a significant drop in performance on SCAN \addjump and CoGnition.
Finally, optimizing our Variational Brown Clustering loss brings extra benefits compared to the previous objective (`-Brown') proposed by~\citet{stratos-2019-mutual}.

\subsection{Analyzing the Embedding Space}
\label{ssec:analysis_embedding}
In this part, we analyze the embedding space learned by \modelname using SoVQ.

\paragraph{Visualizing SCAN embedding space.}
We visualize the embedding matrices using t-distributed Stochastic Neighbor Embedding (t-SNE), which projects each embedding into a 2-dimensional coordinate~\cite{hinton2002stochastic}.
In~\figref{subfig:scan_src_quantized}, we show that the source embeddings learned by SoVQ on SCAN \addjump are clustered based on their syntactic functions in the sentence structure: 
the conjunction words (`\textit{and}', `\textit{after}'), direction adverbs (`\textit{left}', `\textit{right}'), prepositions (`\textit{around}', `\textit{opposite}'), and adverbs (`\textit{twice}', `\textit{thrice}') are clustered together respectively.
Most importantly, the rare primitive `\textit{jump}' is clustered together with other primitives at the top right corner.
This enables the \modelname with SAL to generalize to unseen expressions like ``\textit{jump twice}'' by reusing the same attention pattern as it computes for other expressions (``\textit{walk/look/run twice}'').

On the contrary, in~\figref{subfig:scan_src_baseline}, words of the same syntactic functions (e.g., `\textit{jump}' and `\textit{walk}') are distant apart in the t-SNE space.
This prevents the Transformer from generalizing to novel expressions like ``jump twice''.
We visualize the t-SNE embeddings learned on COGS in~\appendixref{appendixssec:cogs_embed_visualization} and show a similar SoVQ word clustering based on their syntactic functions.

\begin{figure*}[t]
\centering
\subfloat[``\textit{walk around left}'', trained on \addjump.]{\label{subfig:x1_attn_maps_walkaroundleft_app} \includegraphics[width=0.23\textwidth]{figures/x1_attn_maps_walkaroundleft.pdf}}
\hfill
\subfloat[``\textit{jump around left}'', trained on \addjump (KL: 0.337).]{\label{subfig:x1_attn_maps_jumparoundleft_app} \includegraphics[width=0.23\textwidth]{figures/x1_attn_maps_jumparoundleft_v2_thicker.pdf}}%
\hfill
\subfloat[``\textit{walk around left}'', trained on \addjump 2x.]{\label{subfig:x2_attn_maps_walkaroundleft_app} \includegraphics[width=0.23\textwidth]{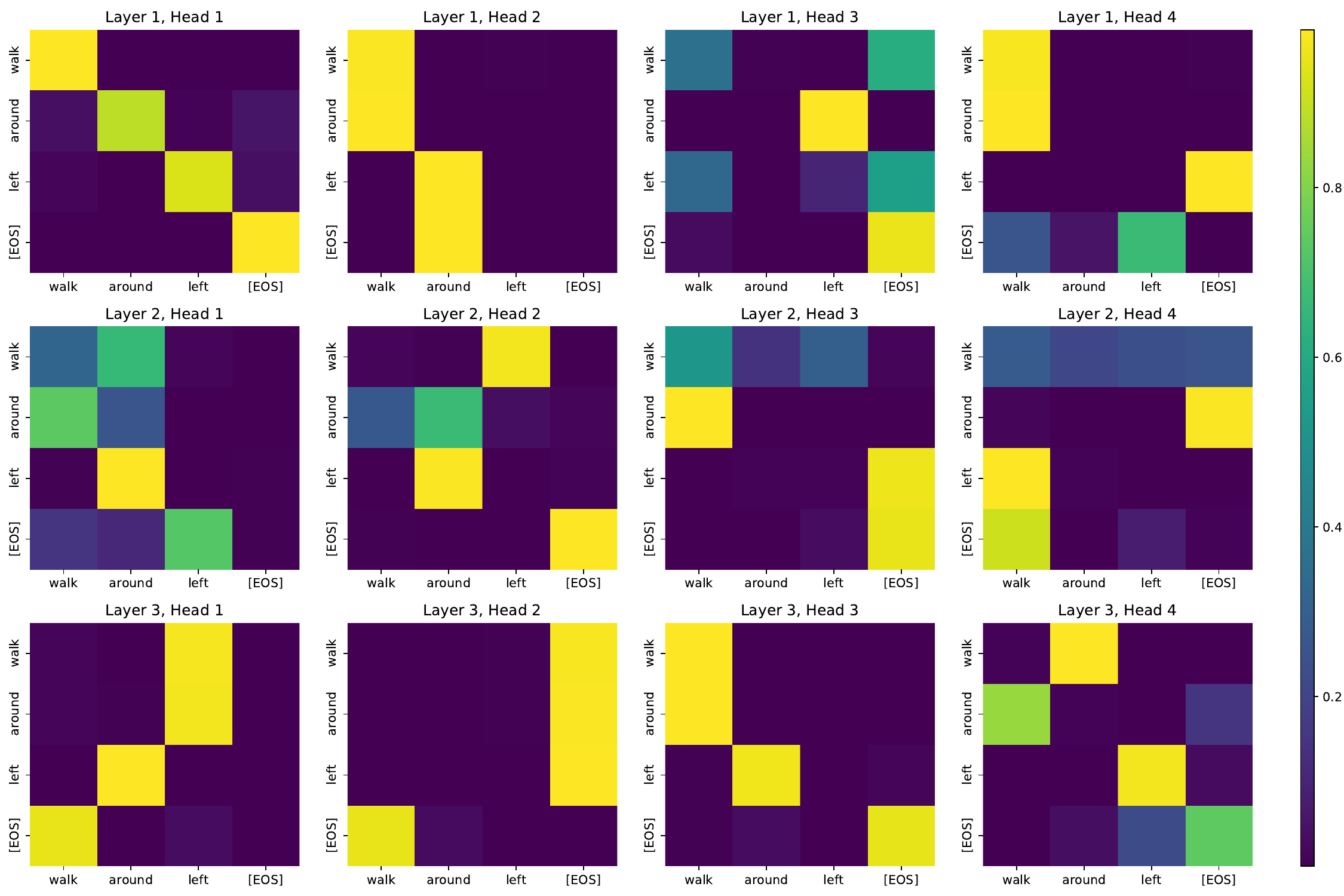}}
\hfill
\subfloat[``\textit{jump around left}'', trained on \addjump 2x (KL: 0.080).]{\label{subfig:x2_attn_maps_jumparoundleft_app} \includegraphics[width=0.23\textwidth]{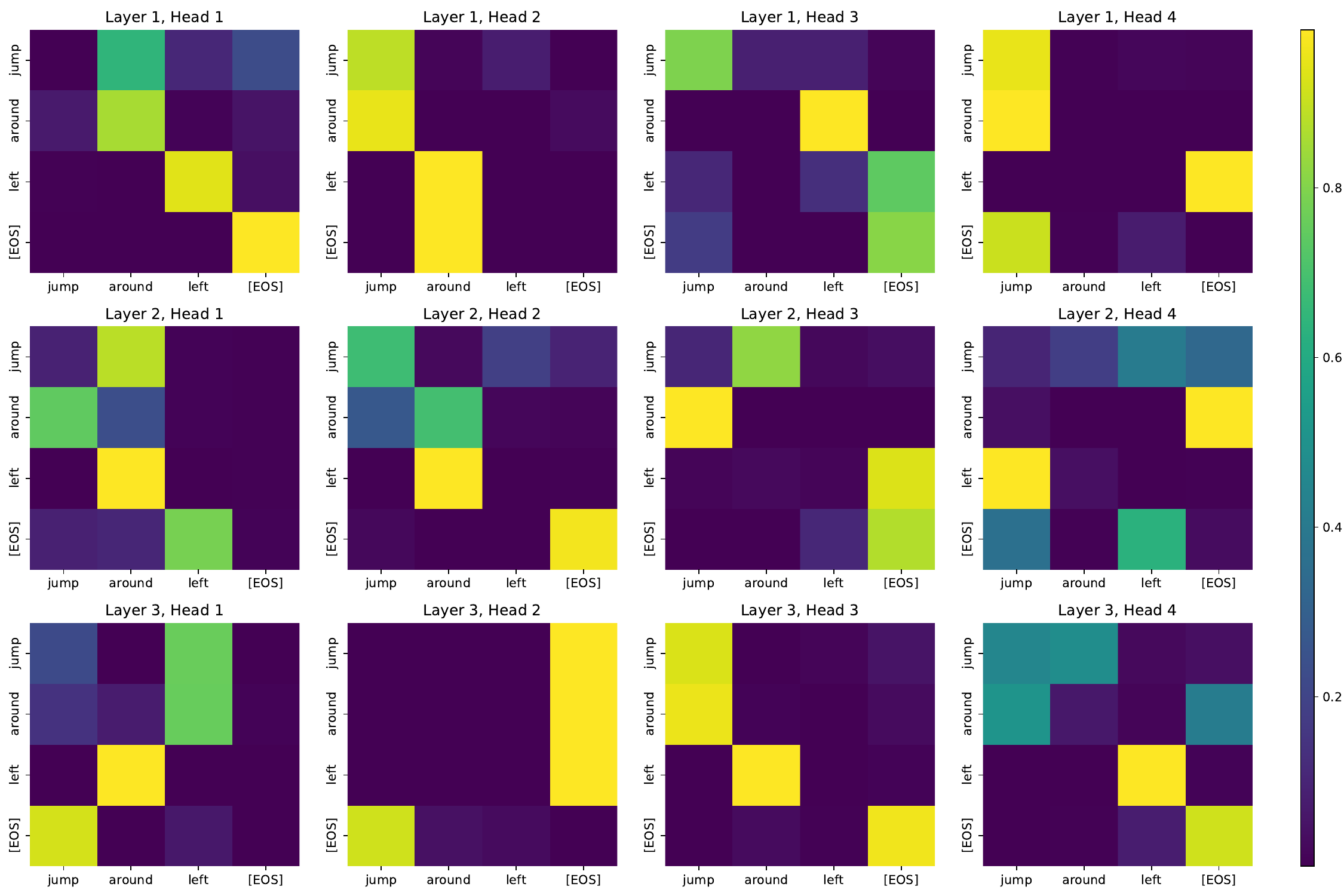}}
  \vspace{0.5em}
\subfloat[``\textit{walk around left}'', trained on \addjump 20x.]{\label{subfig:x20_attn_maps_walkaroundleft_app} \includegraphics[width=0.23\textwidth]{figures/x20_attn_maps_walkaroundleft.pdf}}
\hfill
\subfloat[``\textit{jump around left}'', trained on \addjump 20x (KL:0.001).]{\label{subfig:x20_attn_maps_jumparoundleft_app} \includegraphics[width=0.23\textwidth]{figures/x20_attn_maps_jumparoundleft.pdf}}
\hfill
\subfloat[``\textit{walk around left}'', trained on \addjump 200x.]{\label{subfig:x200_attn_maps_walkaroundleft_app} \includegraphics[width=0.23\textwidth]{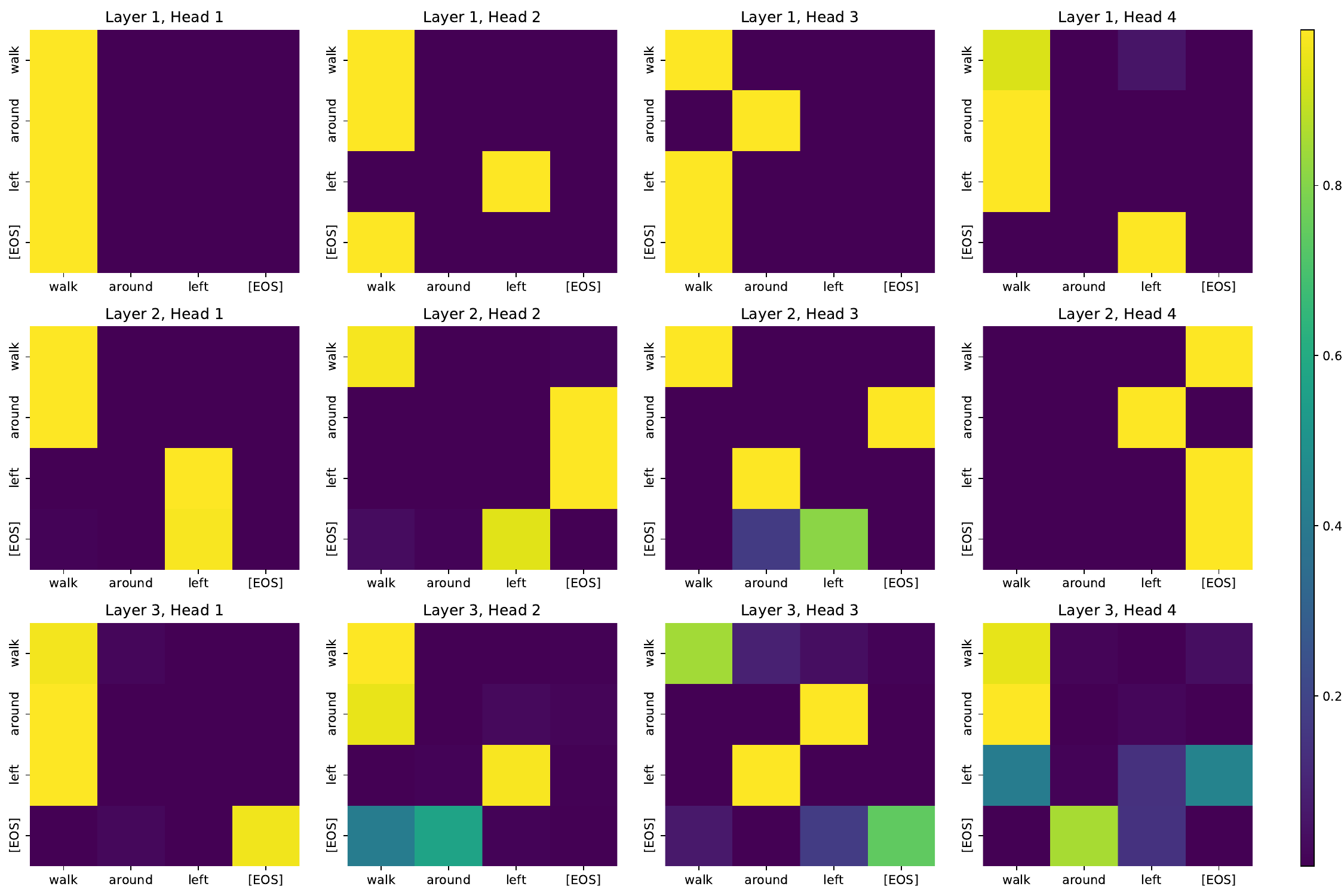}}
\hfill
\subfloat[``\textit{jump around left}'', trained on \addjump 200x (KL:0.001).]{\label{subfig:x200_attn_maps_jumparoundleft_app} \includegraphics[width=0.23\textwidth]{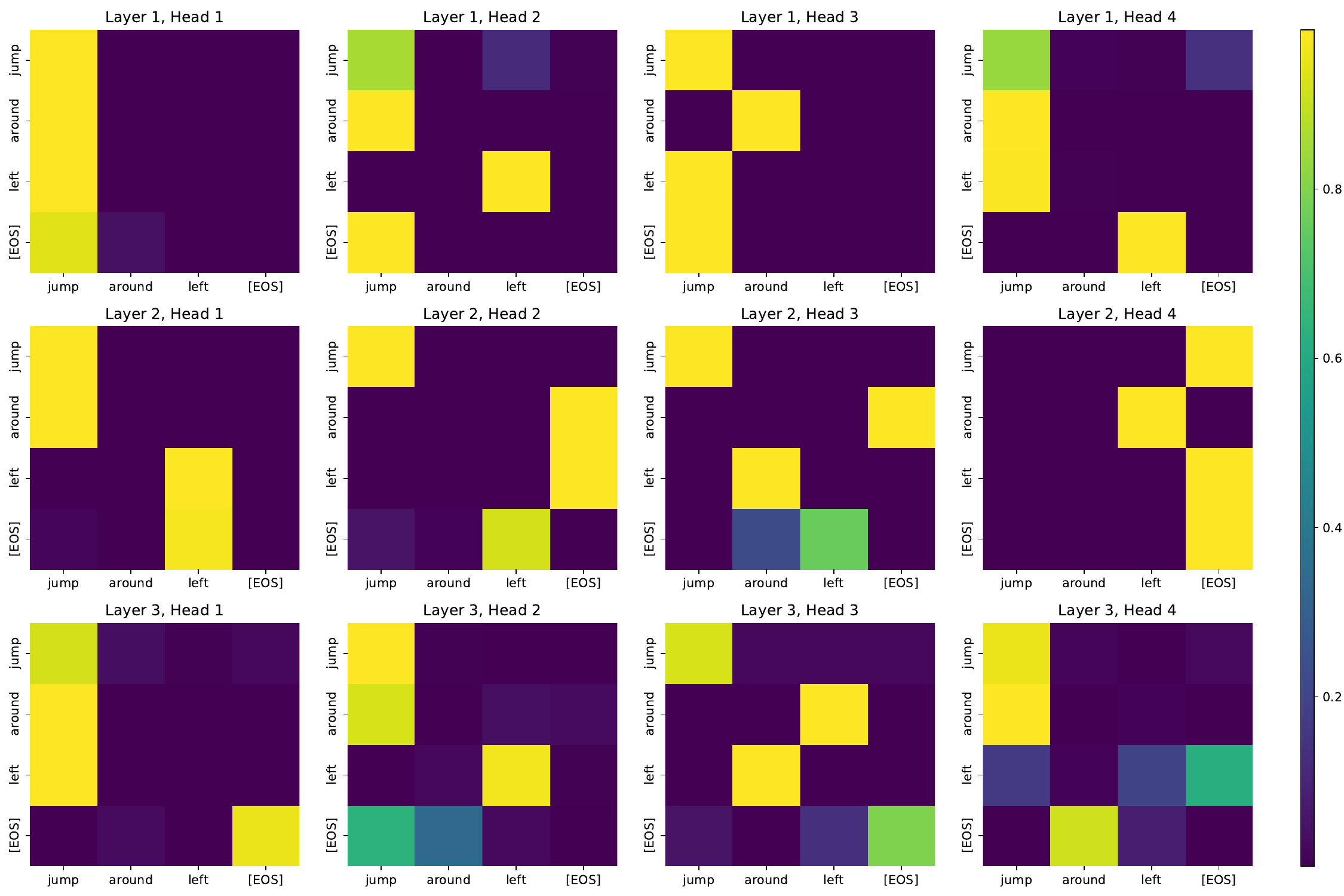}}
\caption{Encoder's attention maps and their average KL-divergence between the two examples, from the vanilla Transformer trained on the SCAN \addjump datasets of different number of distinct primitives~\cite{zhou-jiang-2023-datafactor} (original, 2x, 20x, 200x), where the model achieves 3.67\%, 15.78\%, 100\%, and 100\% accuracy on the entire test set respectively.}
\label{fig:1x-200x_attn_maps}
\end{figure*}

\paragraph{Case Study: Learning the syntactic equivalence of `jump' and `walk'.}
The major challenge in SCAN \addjump is to recognize the equivalent syntactic function of the rare primitive `\textit{jump}' and other common primitives like `\textit{walk}' based on the only syntactic structure\footnote{Atomic expressions like ``\textit{jump}$\mapsto$\texttt{JUMP}'', ``\textit{walk}$\mapsto$\texttt{WALK}.''} that has both `\textit{jump}' and `\textit{walk}' as a constituent.
Next, we present a case study of how SoVQ manages to learn this equivalence and demonstrate three necessary preconditions.
\textbf{First, it is important to choose the proper number of clusters so that the model cannot afford to reserve a cluster for `\textit{jump}' only.}
For example, if we initialize 12 classes instead of 6 classes for the vector quantization, the model will put `\textit{jump}' into a separate class from `\textit{walk}' and `\textit{run}'.
\textbf{Second, the model must be exposed to a sufficient number of examples that put `\textit{jump}' and `\textit{walk}' in the same syntactic structure (context)}.
In~\figref{fig:ablation_num_atomic}, we show the t-SNE visualization of the source embeddings and generalization accuracy when the model is trained with different ($n_a$) repetitions of atomic expressions for each primitive.\footnote{``$n_a$=1000'' means there are 1000 ``\textit{jump}$\mapsto$\texttt{JUMP}'', 1000 ``\textit{walk}$\mapsto$\texttt{WALK}'', etc. in the training set.}
In all four cases, SoVQ can roughly cluster words based on their syntactic functions (e.g., `twice' and `thrice' are always together). 
However, there are some subtle differences regarding the clustering of verb primitives (shown in red).
When $n_a$ equals 1 or 10 (\figref{subfig:scan_src_jump1}, \figref{subfig:scan_src_jump10}), \modelname can hardly generalize (accuracy < 5\%) and the `\textit{jump}' is located distantly with other verbs in red.
With 100 atomic expressions per primitive (\figref{subfig:scan_src_jump100}), \modelname can generalize to 79.38\% of novel expressions, and the verbs are distributed more closely together.
Ultimately, with 1000 atomic expressions per primitive (\figref{subfig:scan_src_jump1000}), \modelname achieves perfect generalization and learns the most compact cluster of verbs.
\textbf{Finally, minimizing the cross-entropy between the cluster inference distribution and cluster prediction distribution is indispensable for clustering words based on their syntactic functions.}
Based on~\theoremref{theo:cross_entropy}, if we optimize this objective on expressions that have `\textit{jump}' and `\textit{walk}' in the same context,
any of its optimal solutions must cluster `\textit{jump}' and `\textit{walk}' into the same class.
Empirically, we find that this theorem can be generalized to a more realistic and noisy setting: 
even when `\textit{walk}' also appears in a lot of other contexts that `\textit{jump}' is never associated with,
SoVQ can still cluster them together based on a small portion of atomic expressions that have both entities in the same context.

\subsection{Analyzing the Attention Patterns}
\label{ssec:analysis_circuits}

\paragraph{Visualizing the attention patterns of vanilla Transformer.}
As is discussed in~\secref{sec:bg}, we train a vanilla Transformer model with 3 layers and 4 heads on the SCAN \addjump training set of different complexities~\cite{zhou-jiang-2023-datafactor}
For example, the 20x augmented training set has 20 times more primitives (excluding jump) that appear in a variety of syntax structures.
In~\figref{fig:1x-200x_attn_maps}, we visualize its attention maps from processing two separate expressions: ``\textit{walk around left}'', which is in the training set, and ``\textit{jump around left}'', which is excluded from the training set.
We also show the generalization accuracy and Kullback–Leibler (KL) divergence between the two attention distributions, averaged over all heads and layers.
In summary, we show that as the number of distinct primitives increases, the KL divergence between the attention distributions of these two examples decreases.
This improvement in attention similarity positively correlates with the improvement of the generalization accuracy.
With 20x more lexical primitives than the original training set, the KL divergence reaches 0.001 and the accuracy reaches 100\%.
This observation provides a mechanistic explanation of how a Transformer trained on more complex data acquires better out-of-domain generalization to novel combinations of structure and lexical entities.

\paragraph{Analyzing the attention patterns of SRL.}
In this part, we reveal that \modelname indeed learns attention patterns that systematically generalize to novel expressions.
As we have shown above, SoVQ can cluster words in the SCAN vocab into multiple structural equivalence classes ${\{\mathcal{C}_i\}}$ where tokens within a class have the same syntactic functions.
Therefore, by computing the queries and keys using the quantized embeddings, Systematic Attention Layers (SAL) are guaranteed to produce the same attention maps given any examples of a common structure (e.g., $\$x$ \textit{around left} $\forall\$x\in \mathcal{C}_i$),
even though an expression like ``\textit{jump around left}'' is never seen in training.
This implies that the model has learned a generalizable attention pattern for representing a common structural relation among classes of equivalently cognizable entities. 

Next, we focus on analyzing the attention maps learned by the Systematically Regularized Layer (SRL).
Unlike SAL, SRL still computes the attention weights using unquantized word embeddings as the queries and keys,
while its outputs are regularized by the attention outputs using the quantized embeddings as the queries, keys, and values (\figref{fig:srl}).
To evaluate its effectiveness, we collect 48 pairs of source sentences from the CoGnition test set.
The sentences within each pair are quantized into the same sequence of clusters (e.g., ``\textit{he stopped every girl.}'' and ``\textit{he found each child.}'') by our \modelname.
We show all 48 pairs of sentences in~\tabref{table:same_code_seq_examples}.

Then, for each pair, we calculate the KL divergence between their attention distributions across all heads and all SRL layers as a metric to evaluate the model's systematicity on unseen sentences.
The \modelname achieves a 0.096 average score, while the baseline Transformer has an infinite KL divergence in 38 out of 48 examples.
We also count the percentage of attention heads that assign the highest weight to the same token when processing two examples in a pair.
In this metric, \modelname achieves 79.8\% compared to the baseline with 72.8\%.
This demonstrates that SRL effectively learns more systematic attention patterns in representing a common structure, while maintaining flexibility in representing non-structural relations, which we will discuss in~\secref{ssec:discussion_compositionality}.

\section{Discussion}
\subsection{Compositionality of Natural Language}
\label{ssec:discussion_compositionality}

In this part, we (1) explain why the Systematic Attention Layer (SAL) can only work on small, synthetic datasets like SCAN;
(2) further motivate the Systematically Regularized Layer (SRL) that achieves a balance between being compositional in encoding common structures and maintaining flexibility in incorporating non-compositional phenomena from larger, more natural datasets.

Recall that \textit{the principle of systematicity involves a capacity to represent/process common structural relations among the equivalently cognizable entities}~\cite{phillips2016systematicity}.
SAL represents the ``common structural relations'' by computing attention weights in a more abstract space, using the quantized word embedding that only encodes the syntactic functions as the queries and keys.
This ensures that sentences that have the same structure and equivalently cognizable entities at all positions (e.g., ``\textit{The cat is awake}'' and ``\textit{The dog is asleep}'') are processed with the same attention weights across all heads and layers.
However, natural languages are only approximately compositional, indicating that sometimes the meaning of a piece of text not only depends on its structure and the meaning of its lexical constituents/entities.
We discuss two situations where having SAL is overly strict and thus prevents the model from encoding certain linguistic features that do not follow the principle of compositionality.

\paragraph{Situation 1: Semantics.}
We use the classic Winograd Challenge~\cite{levesque2012winograd} to demonstrate how lexical semantics, or more specifically, commonsense knowledge affects the comprehension of a sentence. 
Here, the model sees two sentences that differ only in one or two words, which are often of the same syntactic role but contain a referential ambiguity resolved in opposite directions.
For example:
\begin{itemize}
  \item The trophy doesn’t fit in the brown suitcase because it’s
too big. What is too big? \textbf{Answer}: The trophy.
  \item The trophy doesn’t fit in the brown suitcase because it’s
too small. What is too small? \textbf{Answer}: The suitcase.
\end{itemize}
Here, it is necessary to use semantics, rather than syntax only, to resolve the ambiguous anaphora. 
However, the attention maps learned by SAL for these two sentences would be the same, given that ``big'' and ``small'' are most likely quantized into the same class.
Therefore, the model cannot utilize the commonsense knowledge to associate ``big/small'' with ``trophy/suitcase'' by adjusting the attention weights.

\paragraph{Situation 2: Pragmatics.}
Other than the inability to incorporate commonsense to resolve the co-reference within a sentence, SAL with ``hard invariance'' with respect to structural equivalence classes is only designed for capturing dependency between words within a sentence.
It cannot scale to model the complex relationships between words across sentences.
Previously, \citet{sartran-etal-2022-transformer} bias the attention with the sentence parse tree, and reported deteriorated performance on document-level language modeling.
Similarly, we observe that using quantized embeddings to compute cross-attention weights would result in degenerated performance on SCAN tasks, and thus opt for the original word embeddings as the queries and keys (\appendixref{appendix_ssec:sal_decoder}).

In both situations, it is necessary to compute the attention using the word embeddings rather than their quantized code embedding shared among a class of structurally equivalent tokens, so that other non-structural relations (e.g., commonsense) can affect the attention distribution as well.
Empirically, we also show that using the SAL results in degenerated performance on COGS and CoGnition datasets (\tabref{table:ablation}), which include a much larger vocab plus more natural and longer expressions. 
Therefore, as we motivated in~\secref{ssec:method_srl}, we instead use the more flexible SRL to achieve strong performance on these tasks.

\subsection{How does Transformer Generalize Compositionally?}
It has long been argued that neural networks are associative devices that cannot capture systematic compositionality~\cite{fodor1988connectionism,marcus1998rethinking,fodor2002compositionality,marcus2003algebraic,calvo2014architecture}. 
Specifically, \citet{fodor1988connectionism} claimed that ``\textit{in traditional Associationism, the probability that one Idea will elicit another is sensitive to the strength of the association between them. ...
Associative strength was not, however, presumed to be sensitive to features of the content or the structure of representations per se. 
Similarly, in Connectionist models, the selection of an output corresponding to a given input is a function of properties of the paths that connect them.''}
As a result, they further stated that \textit{``\textbf{The syntactic/semantic structure of the representation of an input is not presumed to be a factor in determining the selection of a corresponding output since, as we have seen, syntactic/semantic structure is not defined for the sorts of representations that Connectionist models acknowledge.}''}
After we showcase the systematic behavior of \modelname, readers might ask ``how does our method overcome the inherent limitation of Connectionist models elicited in~\citet{fodor1988connectionism}?''

First, the statement above made an important assumption about neural networks (i.e., Connectionist models) that syntactic/semantic structure does not determine the strength of association between neurons (through the form of attention or full connection).
We agree that the strength of the association is decided by the correlations a neural model observed in data.
Thus, models like Transformers and RNNs fail to execute ``\textit{jump twice}'' because where ``\textit{jump}'' and ``\textit{twice}'' are never seen together in training and models are insensitive to the syntactic structure.

However, we argue that with the proper regularization (e.g., SoVQ and SRL) to the intermediate representations (incl. embeddings and layer outputs), neural models can take the sentence structure into account when determining the strength of inter-neuron association via attention weights or MLP connection.
\textbf{This is because neural models are not inherently limited to capturing shallow, word-level correlations.}
As is shown in~\citet{hewitt-manning-2019-structural}, the attention maps can also be sensitive to the common, although latent, dependency between words, which is simply a kind of statistical correlation between multiple latent and explicit factors (word, position, order, etc).
Therefore, the resulting representations also encode rich structural information. 
The extent to which the model can be sensitive to the structure depends on the data complexity~\cite{zhou-jiang-2023-datafactor}, model architecture~\cite{murty2023pushdown}, and regularization~\cite{jiang-bansal-2021-inducing,yin-etal-2023-consistency}.
For example, there is a statistical correlation between the input word ``\textit{twice}'' and the latent output structure (always repeating the action preceding ``\textit{twice}'' 2 times).
We showed that \modelname can very well capture this association systematically in its attention maps.

\section{Other Related Work}

\paragraph{Compositional generalization.}
Earlier works investigated compositionality of neural networks in language learning~\cite{wong2007generalisation, brakel2009strong}, compositional counting~\cite{wiles1998recurrent,weiss-etal-2018-practical}, and syntax learning~\cite{linzen-etal-2016-assessing}.
Recent works~\cite{lake2018generalization, kim-linzen-2020-cogs,loula-etal-2018-rearranging,bastings-etal-2018-jump,keysers2020measuring,tsarkov2020cfq,hupkes2020compositionality} embed the compositional challenge into semantic parsing tasks and directly evaluate seq2seq models on an out-of-distribution test set.

Previous works have proposed many novel methods to improve the compositional generalization of neural models.
Such methods include novel architectures~\cite{li-etal-2019-compositional,russin2020systematicity,dessi-baroni-2019-cnns,Gordon2020Permutation,oren-etal-2020-improving,zheng-lapata-2021-compositional-generalization}, grammar-based approaches~\cite{shaw-etal-2021-compositional,kim2021sequencetosequence}, task decomposition~\cite{herzig2021unlocking}, data augmentation~\cite{andreas-2020-good,akyurek2021learning,akyurek2022compositionality}, careful architecture selection~\cite{csordas-etal-2021-devil}, and novel learning methods~\cite{lake2019metaseq2seq,conklin-etal-2021-meta,jiang-etal-2022-mutual,xu2022compositional}.

\paragraph{Structures captured by Transformer.}
Researchers have long been studying how the attention maps of Transformers encode the syntactic structure (e.g., dependency parse) of a sentence~\cite{hewitt-manning-2019-structural,phang2019attention,clark-etal-2019-bert,limisiewicz-etal-2020-universal}.
\citet{murty2023characterizing} projected a Transformer into the space of tree-structured models to uncover the intrinsic compositionality.
Recently, \citet{jian2023syntacticsubstitute} substituted words in a sentence with words from the same syntactic category and then averaged the attention maps of a BERT~\cite{devlin-etal-2019-bert} model across these ``syntactically invariant sentences''. 
Our analysis in~\figref{fig:20x_attn_maps} also makes use of such syntactically invariant sentences.
We further reveal the correlation between the emergence of such systematic attention maps and the emergence of the model's generalization ability in a small Transformer trained from scratch (and this leads to our effort in injecting linguistic structure into the model, as discussed below). 

\paragraph{Incorporating linguistic knowledge into models.}
Many previous works have tried to incorporate linguistically-informed labels, especially syntactic labels, into neural networks~\cite{sennrich-haddow-2016-linguistic,strubell-etal-2018-linguistically,sachan-etal-2021-syntax,qian-etal-2021-structural,sartran-etal-2022-transformer}.
Following this idea, later works explored the syntactic equivalence of `\textit{jump}' and other verbs to induce compositionality~\cite{akyurek-andreas-2021-lexicon,jiang-bansal-2021-inducing,white-cotterell-2022-equivariant} for SCAN. 
Most relevantly,~\citet{chakravarthy-etal-2022-systematicity} manually implemented several abstract ``roles'' for tokens in SCAN vocabulary and computed the attention using the role embedding as the keys and queries. 
Our work differs from most of these works in that we do not require any external labels (e.g., parse tree) of sentences and instead automatically infer the syntactic ``roles'' for each token using Structure-oriented Vector Quantization and leverage them in Systematic Attention Layers and Systematically Regularized Layers. 

\section{Conclusion}
In this work, we propose \modelname with Structure-oriented Vector Quantization and two types of attention layers that use the quantized embeddings as the keys and values.
Our experiments show that \modelname can better generalize to unseen expressions in multiple semantic parsing and machine translation tasks.
We conduct multiple analyses and show that SoVQ can cluster word embeddings based on their syntactic roles and the model learns systematic attention patterns in processing sentences of the same syntactic structure.

\section{Limitations}
In this work, the proposed Structure-oriented Vector Quantization mainly clusters the \textit{lexical} constituents based on their syntactic roles.
However, it does not explicitly encourage \textit{phrasal} constituents that have the same syntactic role to be close together in the representation space.
Therefore, we find that \modelname does not achieve better performance than the vanilla Transformer on COGS test examples with ``deeper recursion depth'' or ``novel combination modified phrases and grammatical roles'', both of which require generalizing to novel combinations of grammatical structures and phrasal constituents.
We leave this more challenging generalization problem to future work.

\section*{Acknowledgements}
We thank Elias Stengel-Eskin, Peter Hase, Archiki Prasad, Swarnadeep Saha, and Yi-Lin Sung for their useful feedback.
This work was supported by NSF-CAREER Award 1846185, DARPA MCS Grant N66001-19-2-4031, and an Apple Scholars in AI/ML PhD Fellowship.
The views are those of the authors and not of the funding agency.

\bibliography{custom}

\begin{thebibliography}{78}
\expandafter\ifx\csname natexlab\endcsname\relax\def\natexlab#1{#1}\fi

\bibitem[{Agustsson et~al.(2017)Agustsson, Mentzer, Tschannen, Cavigelli, Timofte, Benini, and Gool}]{agustsson2017soft}
Eirikur Agustsson, Fabian Mentzer, Michael Tschannen, Lukas Cavigelli, Radu Timofte, Luca Benini, and Luc~V Gool. 2017.
\newblock Soft-to-hard vector quantization for end-to-end learning compressible representations.
\newblock \emph{Advances in neural information processing systems}, 30.

\bibitem[{Aky{\"u}rek et~al.(2021)Aky{\"u}rek, Aky{\"u}rek, and Andreas}]{akyurek2021learning}
Ekin Aky{\"u}rek, Afra~Feyza Aky{\"u}rek, and Jacob Andreas. 2021.
\newblock \href {https://openreview.net/forum?id=PS3IMnScugk} {Learning to recombine and resample data for compositional generalization}.
\newblock In \emph{International Conference on Learning Representations}.

\bibitem[{Akyurek and Andreas(2021)}]{akyurek-andreas-2021-lexicon}
Ekin Akyurek and Jacob Andreas. 2021.
\newblock \href {https://doi.org/10.18653/v1/2021.acl-long.382} {Lexicon learning for few shot sequence modeling}.
\newblock In \emph{Proceedings of the 59th Annual Meeting of the Association for Computational Linguistics and the 11th International Joint Conference on Natural Language Processing (Volume 1: Long Papers)}, pages 4934--4946, Online. Association for Computational Linguistics.

\bibitem[{Aky{\"u}rek and Andreas(2022)}]{akyurek2022compositionality}
Ekin Aky{\"u}rek and Jacob Andreas. 2022.
\newblock Compositionality as lexical symmetry.
\newblock \emph{arXiv preprint arXiv:2201.12926}.

\bibitem[{Andreas(2020)}]{andreas-2020-good}
Jacob Andreas. 2020.
\newblock \href {https://doi.org/10.18653/v1/2020.acl-main.676} {Good-enough compositional data augmentation}.
\newblock In \emph{Proceedings of the 58th Annual Meeting of the Association for Computational Linguistics}, pages 7556--7566, Online. Association for Computational Linguistics.

\bibitem[{Bastings et~al.(2018)Bastings, Baroni, Weston, Cho, and Kiela}]{bastings-etal-2018-jump}
Jasmijn Bastings, Marco Baroni, Jason Weston, Kyunghyun Cho, and Douwe Kiela. 2018.
\newblock \href {https://doi.org/10.18653/v1/W18-5407} {Jump to better conclusions: {SCAN} both left and right}.
\newblock In \emph{Proceedings of the 2018 {EMNLP} Workshop {B}lackbox{NLP}: Analyzing and Interpreting Neural Networks for {NLP}}, pages 47--55, Brussels, Belgium. Association for Computational Linguistics.

\bibitem[{Bojar et~al.(2017)Bojar, Chatterjee, Federmann, Graham, Haddow, Huang, Huck, Koehn, Liu, Logacheva, Monz, Negri, Post, Rubino, Specia, and Turchi}]{wmt17}
Ond~{r}ej Bojar, Rajen Chatterjee, Christian Federmann, Yvette Graham, Barry Haddow, Shujian Huang, Matthias Huck, Philipp Koehn, Qun Liu, Varvara Logacheva, Christof Monz, Matteo Negri, Matt Post, Raphael Rubino, Lucia Specia, and Marco Turchi. 2017.
\newblock \href {http://www.aclweb.org/anthology/W17-4717} {Findings of the 2017 conference on machine translation (wmt17)}.
\newblock In \emph{Proceedings of the Second Conference on Machine Translation, Volume 2: Shared Task Papers}, pages 169--214, Copenhagen, Denmark. Association for Computational Linguistics.

\bibitem[{Bojar et~al.(2014)Bojar, Buck, Federmann, Haddow, Koehn, Leveling, Monz, Pecina, Post, Saint-Amand, Soricut, Specia, and Tamchyna}]{wmt14}
Ondrej Bojar, Christian Buck, Christian Federmann, Barry Haddow, Philipp Koehn, Johannes Leveling, Christof Monz, Pavel Pecina, Matt Post, Herve Saint-Amand, Radu Soricut, Lucia Specia, and Ale\v{s} Tamchyna. 2014.
\newblock \href {http://www.aclweb.org/anthology/W/W14/W14-3302} {Findings of the 2014 workshop on statistical machine translation}.
\newblock In \emph{Proceedings of the Ninth Workshop on Statistical Machine Translation}, pages 12--58, Baltimore, Maryland, USA. Association for Computational Linguistics.

\bibitem[{Brakel and Frank(2009)}]{brakel2009strong}
Phil{\'e}mon Brakel and Stefan Frank. 2009.
\newblock Strong systematicity in sentence processing by simple recurrent networks.
\newblock In \emph{Proceedings of the Annual Meeting of the Cognitive Science Society}, volume~31.

\bibitem[{Brown et~al.(1992)Brown, Della~Pietra, Desouza, Lai, and Mercer}]{brown1992class}
Peter~F Brown, Vincent~J Della~Pietra, Peter~V Desouza, Jennifer~C Lai, and Robert~L Mercer. 1992.
\newblock Class-based n-gram models of natural language.
\newblock \emph{Computational linguistics}, 18(4):467--480.

\bibitem[{Calvo and Symons(2014)}]{calvo2014architecture}
Paco Calvo and John Symons. 2014.
\newblock \emph{The Architecture of Cognition: Rethinking Fodor and Pylyshyn's Systematicity Challenge}.
\newblock MIT Press.

\bibitem[{Chakravarthy et~al.(2022)Chakravarthy, Russin, and O{'}Reilly}]{chakravarthy-etal-2022-systematicity}
Ayush~K Chakravarthy, Jacob~Labe Russin, and Randall O{'}Reilly. 2022.
\newblock \href {https://doi.org/10.18653/v1/2022.naacl-srw.1} {Systematicity emerges in transformers when abstract grammatical roles guide attention}.
\newblock In \emph{Proceedings of the 2022 Conference of the North American Chapter of the Association for Computational Linguistics: Human Language Technologies: Student Research Workshop}, pages 1--8, Hybrid: Seattle, Washington + Online. Association for Computational Linguistics.

\bibitem[{Chomsky(1957)}]{chomsky1957syntactic}
Noam Chomsky. 1957.
\newblock \emph{Syntactic structures}.
\newblock De Gruyter Mouton.

\bibitem[{Clark et~al.(2019)Clark, Khandelwal, Levy, and Manning}]{clark-etal-2019-bert}
Kevin Clark, Urvashi Khandelwal, Omer Levy, and Christopher~D. Manning. 2019.
\newblock \href {https://doi.org/10.18653/v1/W19-4828} {What does {BERT} look at? an analysis of {BERT}{'}s attention}.
\newblock In \emph{Proceedings of the 2019 ACL Workshop BlackboxNLP: Analyzing and Interpreting Neural Networks for NLP}, pages 276--286, Florence, Italy. Association for Computational Linguistics.

\bibitem[{Conklin et~al.(2021)Conklin, Wang, Smith, and Titov}]{conklin-etal-2021-meta}
Henry Conklin, Bailin Wang, Kenny Smith, and Ivan Titov. 2021.
\newblock \href {https://doi.org/10.18653/v1/2021.acl-long.258} {Meta-learning to compositionally generalize}.
\newblock In \emph{Proceedings of the 59th Annual Meeting of the Association for Computational Linguistics and the 11th International Joint Conference on Natural Language Processing (Volume 1: Long Papers)}, pages 3322--3335, Online. Association for Computational Linguistics.

\bibitem[{Csord{\'a}s et~al.(2021)Csord{\'a}s, Irie, and Schmidhuber}]{csordas-etal-2021-devil}
R{\'o}bert Csord{\'a}s, Kazuki Irie, and Juergen Schmidhuber. 2021.
\newblock \href {https://doi.org/10.18653/v1/2021.emnlp-main.49} {The devil is in the detail: Simple tricks improve systematic generalization of transformers}.
\newblock In \emph{Proceedings of the 2021 Conference on Empirical Methods in Natural Language Processing}, pages 619--634, Online and Punta Cana, Dominican Republic. Association for Computational Linguistics.

\bibitem[{Dess{\`\i} and Baroni(2019)}]{dessi-baroni-2019-cnns}
Roberto Dess{\`\i} and Marco Baroni. 2019.
\newblock \href {https://doi.org/10.18653/v1/P19-1381} {{CNN}s found to jump around more skillfully than {RNN}s: Compositional generalization in seq2seq convolutional networks}.
\newblock In \emph{Proceedings of the 57th Annual Meeting of the Association for Computational Linguistics}, pages 3919--3923, Florence, Italy. Association for Computational Linguistics.

\bibitem[{Devlin et~al.(2019)Devlin, Chang, Lee, and Toutanova}]{devlin-etal-2019-bert}
Jacob Devlin, Ming-Wei Chang, Kenton Lee, and Kristina Toutanova. 2019.
\newblock \href {https://doi.org/10.18653/v1/N19-1423} {{BERT}: Pre-training of deep bidirectional transformers for language understanding}.
\newblock In \emph{Proceedings of the 2019 Conference of the North {A}merican Chapter of the Association for Computational Linguistics: Human Language Technologies, Volume 1 (Long and Short Papers)}, pages 4171--4186, Minneapolis, Minnesota. Association for Computational Linguistics.

\bibitem[{Drozdov et~al.(2023)Drozdov, Sch{\"a}rli, Aky{\"u}rek, Scales, Song, Chen, Bousquet, and Zhou}]{drozdov2023compositional}
Andrew Drozdov, Nathanael Sch{\"a}rli, Ekin Aky{\"u}rek, Nathan Scales, Xinying Song, Xinyun Chen, Olivier Bousquet, and Denny Zhou. 2023.
\newblock \href {https://openreview.net/forum?id=gJW8hSGBys8} {Compositional semantic parsing with large language models}.
\newblock In \emph{The Eleventh International Conference on Learning Representations}.

\bibitem[{Fodor and Lepore(2002)}]{fodor2002compositionality}
Jerry~A Fodor and Ernest Lepore. 2002.
\newblock \emph{The compositionality papers}.
\newblock Oxford University Press.

\bibitem[{Fodor and Pylyshyn(1988)}]{fodor1988connectionism}
Jerry~A Fodor and Zenon~W Pylyshyn. 1988.
\newblock Connectionism and cognitive architecture: A critical analysis.
\newblock \emph{Cognition}, 28(1-2):3--71.

\bibitem[{Furrer et~al.(2020)Furrer, van Zee, Scales, and Sch{\"a}rli}]{furrer2020compositional}
Daniel Furrer, Marc van Zee, Nathan Scales, and Nathanael Sch{\"a}rli. 2020.
\newblock Compositional generalization in semantic parsing: Pre-training vs. specialized architectures.
\newblock \emph{arXiv preprint arXiv:2007.08970}.

\bibitem[{Gordon et~al.(2020)Gordon, Lopez-Paz, Baroni, and Bouchacourt}]{Gordon2020Permutation}
Jonathan Gordon, David Lopez-Paz, Marco Baroni, and Diane Bouchacourt. 2020.
\newblock \href {https://openreview.net/forum?id=SylVNerFvr} {Permutation equivariant models for compositional generalization in language}.
\newblock In \emph{International Conference on Learning Representations}.

\bibitem[{Herzig et~al.(2021)Herzig, Shaw, Chang, Guu, Pasupat, and Zhang}]{herzig2021unlocking}
Jonathan Herzig, Peter Shaw, Ming-Wei Chang, Kelvin Guu, Panupong Pasupat, and Yuan Zhang. 2021.
\newblock Unlocking compositional generalization in pre-trained models using intermediate representations.
\newblock \emph{arXiv preprint arXiv:2104.07478}.

\bibitem[{Hewitt and Manning(2019)}]{hewitt-manning-2019-structural}
John Hewitt and Christopher~D. Manning. 2019.
\newblock \href {https://doi.org/10.18653/v1/N19-1419} {{A} structural probe for finding syntax in word representations}.
\newblock In \emph{Proceedings of the 2019 Conference of the North {A}merican Chapter of the Association for Computational Linguistics: Human Language Technologies, Volume 1 (Long and Short Papers)}, pages 4129--4138, Minneapolis, Minnesota. Association for Computational Linguistics.

\bibitem[{Hinton and Roweis(2002)}]{hinton2002stochastic}
Geoffrey~E Hinton and Sam Roweis. 2002.
\newblock Stochastic neighbor embedding.
\newblock \emph{Advances in neural information processing systems}, 15.

\bibitem[{Hupkes et~al.(2020)Hupkes, Dankers, Mul, and Bruni}]{hupkes2020compositionality}
Dieuwke Hupkes, Verna Dankers, Mathijs Mul, and Elia Bruni. 2020.
\newblock Compositionality decomposed: How do neural networks generalise?
\newblock \emph{Journal of Artificial Intelligence Research}, 67:757--795.

\bibitem[{Jian and Reddy(2023)}]{jian2023syntacticsubstitute}
Jasper Jian and Siva Reddy. 2023.
\newblock Syntactic substitutability as unsupervised dependency syntax.
\newblock In \emph{Proceedings of EMNLP}.

\bibitem[{Jiang and Bansal(2021)}]{jiang-bansal-2021-inducing}
Yichen Jiang and Mohit Bansal. 2021.
\newblock \href {https://doi.org/10.18653/v1/2021.emnlp-main.505} {Inducing transformer{'}s compositional generalization ability via auxiliary sequence prediction tasks}.
\newblock In \emph{Proceedings of the 2021 Conference on Empirical Methods in Natural Language Processing}, pages 6253--6265, Online and Punta Cana, Dominican Republic. Association for Computational Linguistics.

\bibitem[{Jiang et~al.(2022)Jiang, Zhou, and Bansal}]{jiang-etal-2022-mutual}
Yichen Jiang, Xiang Zhou, and Mohit Bansal. 2022.
\newblock \href {https://doi.org/10.18653/v1/2022.emnlp-main.808} {Mutual exclusivity training and primitive augmentation to induce compositionality}.
\newblock In \emph{Proceedings of the 2022 Conference on Empirical Methods in Natural Language Processing}, pages 11778--11793, Abu Dhabi, United Arab Emirates. Association for Computational Linguistics.

\bibitem[{Johnson(2004)}]{johnson2004systematicity}
Kent Johnson. 2004.
\newblock On the systematicity of language and thought.
\newblock \emph{The Journal of Philosophy}, 101(3):111--139.

\bibitem[{Keysers et~al.(2020)Keysers, Sch{\"a}rli, Scales, Buisman, Furrer, Kashubin, Momchev, Sinopalnikov, Stafiniak, Tihon, Tsarkov, Wang, van Zee, and Bousquet}]{keysers2020measuring}
Daniel Keysers, Nathanael Sch{\"a}rli, Nathan Scales, Hylke Buisman, Daniel Furrer, Sergii Kashubin, Nikola Momchev, Danila Sinopalnikov, Lukasz Stafiniak, Tibor Tihon, Dmitry Tsarkov, Xiao Wang, Marc van Zee, and Olivier Bousquet. 2020.
\newblock \href {https://openreview.net/forum?id=SygcCnNKwr} {Measuring compositional generalization: A comprehensive method on realistic data}.
\newblock In \emph{International Conference on Learning Representations}.

\bibitem[{Kim and Linzen(2020)}]{kim-linzen-2020-cogs}
Najoung Kim and Tal Linzen. 2020.
\newblock \href {https://doi.org/10.18653/v1/2020.emnlp-main.731} {{COGS}: A compositional generalization challenge based on semantic interpretation}.
\newblock In \emph{Proceedings of the 2020 Conference on Empirical Methods in Natural Language Processing (EMNLP)}, pages 9087--9105, Online. Association for Computational Linguistics.

\bibitem[{Kim(2021)}]{kim2021sequencetosequence}
Yoon Kim. 2021.
\newblock \href {https://proceedings.neurips.cc/paper/2021/hash/dd17e652cd2a08fdb8bf7f68e2ad3814-Abstract.html} {Sequence-to-sequence learning with latent neural grammars}.
\newblock In \emph{Advances in Neural Information Processing Systems 34: Annual Conference on Neural Information Processing Systems 2021, NeurIPS 2021, December 6-14, 2021, virtual}, pages 26302--26317.

\bibitem[{Kingma and Ba(2015)}]{kingma15adam}
Diederik~P. Kingma and Jimmy Ba. 2015.
\newblock \href {http://arxiv.org/abs/1412.6980} {Adam: A method for stochastic optimization}.
\newblock In \emph{ICLR (Poster)}.

\bibitem[{Lake and Baroni(2018)}]{lake2018generalization}
Brenden Lake and Marco Baroni. 2018.
\newblock Generalization without systematicity: On the compositional skills of sequence-to-sequence recurrent networks.
\newblock In \emph{International Conference on Machine Learning}, pages 2873--2882. PMLR.

\bibitem[{Lake(2019)}]{lake2019metaseq2seq}
Brenden~M Lake. 2019.
\newblock \href {https://proceedings.neurips.cc/paper/2019/file/f4d0e2e7fc057a58f7ca4a391f01940a-Paper.pdf} {Compositional generalization through meta sequence-to-sequence learning}.
\newblock In \emph{Advances in Neural Information Processing Systems}, volume~32. Curran Associates, Inc.

\bibitem[{Levesque et~al.(2012)Levesque, Davis, and Morgenstern}]{levesque2012winograd}
Hector Levesque, Ernest Davis, and Leora Morgenstern. 2012.
\newblock The winograd schema challenge.
\newblock In \emph{Thirteenth international conference on the principles of knowledge representation and reasoning}.

\bibitem[{Li et~al.(2021)Li, Yin, Chen, and Zhang}]{li-etal-2021-compositional}
Yafu Li, Yongjing Yin, Yulong Chen, and Yue Zhang. 2021.
\newblock \href {https://doi.org/10.18653/v1/2021.acl-long.368} {On compositional generalization of neural machine translation}.
\newblock In \emph{Proceedings of the 59th Annual Meeting of the Association for Computational Linguistics and the 11th International Joint Conference on Natural Language Processing (Volume 1: Long Papers)}, pages 4767--4780, Online. Association for Computational Linguistics.

\bibitem[{Li et~al.(2019)Li, Zhao, Wang, and Hestness}]{li-etal-2019-compositional}
Yuanpeng Li, Liang Zhao, Jianyu Wang, and Joel Hestness. 2019.
\newblock \href {https://doi.org/10.18653/v1/D19-1438} {Compositional generalization for primitive substitutions}.
\newblock In \emph{Proceedings of the 2019 Conference on Empirical Methods in Natural Language Processing and the 9th International Joint Conference on Natural Language Processing (EMNLP-IJCNLP)}, pages 4293--4302, Hong Kong, China. Association for Computational Linguistics.

\bibitem[{Limisiewicz et~al.(2020)Limisiewicz, Mare{\v{c}}ek, and Rosa}]{limisiewicz-etal-2020-universal}
Tomasz Limisiewicz, David Mare{\v{c}}ek, and Rudolf Rosa. 2020.
\newblock \href {https://doi.org/10.18653/v1/2020.findings-emnlp.245} {{U}niversal {D}ependencies {A}ccording to {BERT}: {B}oth {M}ore {S}pecific and {M}ore {G}eneral}.
\newblock In \emph{Findings of the Association for Computational Linguistics: EMNLP 2020}, pages 2710--2722, Online. Association for Computational Linguistics.

\bibitem[{Lingle(2023)}]{lingle2023transformer}
Lucas~D Lingle. 2023.
\newblock Transformer-vq: Linear-time transformers via vector quantization.
\newblock \emph{arXiv preprint arXiv:2309.16354}.

\bibitem[{Linzen et~al.(2016)Linzen, Dupoux, and Goldberg}]{linzen-etal-2016-assessing}
Tal Linzen, Emmanuel Dupoux, and Yoav Goldberg. 2016.
\newblock \href {https://doi.org/10.1162/tacl_a_00115} {Assessing the ability of {LSTM}s to learn syntax-sensitive dependencies}.
\newblock \emph{Transactions of the Association for Computational Linguistics}, 4:521--535.

\bibitem[{Loula et~al.(2018)Loula, Baroni, and Lake}]{loula-etal-2018-rearranging}
Jo{\~a}o Loula, Marco Baroni, and Brenden Lake. 2018.
\newblock \href {https://doi.org/10.18653/v1/W18-5413} {Rearranging the familiar: Testing compositional generalization in recurrent networks}.
\newblock In \emph{Proceedings of the 2018 {EMNLP} Workshop {B}lackbox{NLP}: Analyzing and Interpreting Neural Networks for {NLP}}, pages 108--114, Brussels, Belgium. Association for Computational Linguistics.

\bibitem[{Marcus(1998)}]{marcus1998rethinking}
Gary~F Marcus. 1998.
\newblock Rethinking eliminative connectionism.
\newblock \emph{Cognitive psychology}, 37(3):243--282.

\bibitem[{Marcus(2003)}]{marcus2003algebraic}
Gary~F Marcus. 2003.
\newblock \emph{The algebraic mind: Integrating connectionism and cognitive science}.
\newblock MIT press.

\bibitem[{McAllester(2018)}]{mcallester2018information}
David McAllester. 2018.
\newblock Information theoretic co-training.
\newblock \emph{arXiv preprint arXiv:1802.07572}.

\bibitem[{Montague(1970)}]{montague1970universal}
Richard Montague. 1970.
\newblock Universal grammar.
\newblock \emph{1974}, pages 222--46.

\bibitem[{Murty et~al.(2023{\natexlab{a}})Murty, Sharma, Andreas, and Manning}]{murty2023characterizing}
Shikhar Murty, Pratyusha Sharma, Jacob Andreas, and Christopher~D Manning. 2023{\natexlab{a}}.
\newblock \href {https://openreview.net/forum?id=sAOOeI878Ns} {Characterizing intrinsic compositionality in transformers with tree projections}.
\newblock In \emph{The Eleventh International Conference on Learning Representations}.

\bibitem[{Murty et~al.(2023{\natexlab{b}})Murty, Sharma, Andreas, and Manning}]{murty2023pushdown}
Shikhar Murty, Pratyusha Sharma, Jacob Andreas, and Christopher~D Manning. 2023{\natexlab{b}}.
\newblock Pushdown layers: Encoding recursive structure in transformer language models.
\newblock \emph{arXiv preprint arXiv:2310.19089}.

\bibitem[{Oren et~al.(2020)Oren, Herzig, Gupta, Gardner, and Berant}]{oren-etal-2020-improving}
Inbar Oren, Jonathan Herzig, Nitish Gupta, Matt Gardner, and Jonathan Berant. 2020.
\newblock \href {https://doi.org/10.18653/v1/2020.findings-emnlp.225} {Improving compositional generalization in semantic parsing}.
\newblock In \emph{Findings of the Association for Computational Linguistics: EMNLP 2020}, pages 2482--2495, Online. Association for Computational Linguistics.

\bibitem[{Papineni et~al.(2002)Papineni, Roukos, Ward, and Zhu}]{papineni2002bleu}
Kishore Papineni, Salim Roukos, Todd Ward, and Wei-Jing Zhu. 2002.
\newblock Bleu: a method for automatic evaluation of machine translation.
\newblock In \emph{Proceedings of the 40th annual meeting of the Association for Computational Linguistics}, pages 311--318.

\bibitem[{Phang et~al.(2019)Phang, Bordia, Bowman et~al.}]{phang2019attention}
Jason Phang, Shikha Bordia, Samuel~R Bowman, et~al. 2019.
\newblock Do attention heads in bert track syntactic dependencies?
\newblock In \emph{NY Academy of Sciences NLP, Dialog, and Speech Workshop}.

\bibitem[{Phillips and Wilson(2016)}]{phillips2016systematicity}
Steven Phillips and William~H Wilson. 2016.
\newblock Systematicity and a categorical theory of cognitive architecture: universal construction in context.
\newblock \emph{Frontiers in psychology}, 7:1139.

\bibitem[{Qian et~al.(2021)Qian, Naseem, Levy, and Fernandez~Astudillo}]{qian-etal-2021-structural}
Peng Qian, Tahira Naseem, Roger Levy, and Ram{\'o}n Fernandez~Astudillo. 2021.
\newblock \href {https://doi.org/10.18653/v1/2021.acl-long.289} {Structural guidance for transformer language models}.
\newblock In \emph{Proceedings of the 59th Annual Meeting of the Association for Computational Linguistics and the 11th International Joint Conference on Natural Language Processing (Volume 1: Long Papers)}, pages 3735--3745, Online. Association for Computational Linguistics.

\bibitem[{Ramesh et~al.(2021)Ramesh, Pavlov, Goh, Gray, Voss, Radford, Chen, and Sutskever}]{ramesh2021dalle}
Aditya Ramesh, Mikhail Pavlov, Gabriel Goh, Scott Gray, Chelsea Voss, Alec Radford, Mark Chen, and Ilya Sutskever. 2021.
\newblock Zero-shot text-to-image generation.
\newblock In \emph{International Conference on Machine Learning}, pages 8821--8831. PMLR.

\bibitem[{Razavi et~al.(2019)Razavi, Van~den Oord, and Vinyals}]{razavi2019vqvae2}
Ali Razavi, Aaron Van~den Oord, and Oriol Vinyals. 2019.
\newblock Generating diverse high-fidelity images with vq-vae-2.
\newblock \emph{Advances in neural information processing systems}, 32.

\bibitem[{Russin et~al.(2020)Russin, Jo, O'Reilly, and Bengio}]{russin2020systematicity}
Jacob~L Russin, Jason Jo, Randall~C O'Reilly, and Yoshua Bengio. 2020.
\newblock Systematicity in a recurrent neural network by factorizing syntax and semantics.
\newblock In \emph{CogSci}.

\bibitem[{Sachan et~al.(2021)Sachan, Zhang, Qi, and Hamilton}]{sachan-etal-2021-syntax}
Devendra Sachan, Yuhao Zhang, Peng Qi, and William~L. Hamilton. 2021.
\newblock \href {https://doi.org/10.18653/v1/2021.eacl-main.228} {Do syntax trees help pre-trained transformers extract information?}
\newblock In \emph{Proceedings of the 16th Conference of the European Chapter of the Association for Computational Linguistics: Main Volume}, pages 2647--2661, Online. Association for Computational Linguistics.

\bibitem[{Sartran et~al.(2022)Sartran, Barrett, Kuncoro, Stanojevi{\'c}, Blunsom, and Dyer}]{sartran-etal-2022-transformer}
Laurent Sartran, Samuel Barrett, Adhiguna Kuncoro, Milo{\v{s}} Stanojevi{\'c}, Phil Blunsom, and Chris Dyer. 2022.
\newblock \href {https://doi.org/10.1162/tacl_a_00526} {Transformer grammars: Augmenting transformer language models with syntactic inductive biases at scale}.
\newblock \emph{Transactions of the Association for Computational Linguistics}, 10:1423--1439.

\bibitem[{Sennrich and Haddow(2016)}]{sennrich-haddow-2016-linguistic}
Rico Sennrich and Barry Haddow. 2016.
\newblock \href {https://doi.org/10.18653/v1/W16-2209} {Linguistic input features improve neural machine translation}.
\newblock In \emph{Proceedings of the First Conference on Machine Translation: Volume 1, Research Papers}, pages 83--91, Berlin, Germany. Association for Computational Linguistics.

\bibitem[{Shaw et~al.(2021)Shaw, Chang, Pasupat, and Toutanova}]{shaw-etal-2021-compositional}
Peter Shaw, Ming-Wei Chang, Panupong Pasupat, and Kristina Toutanova. 2021.
\newblock \href {https://doi.org/10.18653/v1/2021.acl-long.75} {Compositional generalization and natural language variation: Can a semantic parsing approach handle both?}
\newblock In \emph{Proceedings of the 59th Annual Meeting of the Association for Computational Linguistics and the 11th International Joint Conference on Natural Language Processing (Volume 1: Long Papers)}, pages 922--938, Online. Association for Computational Linguistics.

\bibitem[{Stratos(2019)}]{stratos-2019-mutual}
Karl Stratos. 2019.
\newblock \href {https://doi.org/10.18653/v1/N19-1113} {Mutual information maximization for simple and accurate part-of-speech induction}.
\newblock In \emph{Proceedings of the 2019 Conference of the North {A}merican Chapter of the Association for Computational Linguistics: Human Language Technologies, Volume 1 (Long and Short Papers)}, pages 1095--1104, Minneapolis, Minnesota. Association for Computational Linguistics.

\bibitem[{Strubell et~al.(2018)Strubell, Verga, Andor, Weiss, and McCallum}]{strubell-etal-2018-linguistically}
Emma Strubell, Patrick Verga, Daniel Andor, David Weiss, and Andrew McCallum. 2018.
\newblock \href {https://doi.org/10.18653/v1/D18-1548} {Linguistically-informed self-attention for semantic role labeling}.
\newblock In \emph{Proceedings of the 2018 Conference on Empirical Methods in Natural Language Processing}, pages 5027--5038, Brussels, Belgium. Association for Computational Linguistics.

\bibitem[{Tishby and Zaslavsky(2015)}]{tishby2015deep}
Naftali Tishby and Noga Zaslavsky. 2015.
\newblock Deep learning and the information bottleneck principle.
\newblock In \emph{2015 ieee information theory workshop (itw)}, pages 1--5. IEEE.

\bibitem[{Tsarkov et~al.(2021)Tsarkov, Tihon, Scales, Momchev, Sinopalnikov, and Sch{\"a}rli}]{tsarkov2020cfq}
Dmitry Tsarkov, Tibor Tihon, Nathan Scales, Nikola Momchev, Danila Sinopalnikov, and Nathanael Sch{\"a}rli. 2021.
\newblock *-cfq: Analyzing the scalability of machine learning on a compositional task.
\newblock In \emph{AAAI}.

\bibitem[{Van Den~Oord et~al.(2017)Van Den~Oord, Vinyals et~al.}]{van2017vqvae}
Aaron Van Den~Oord, Oriol Vinyals, et~al. 2017.
\newblock Neural discrete representation learning.
\newblock \emph{Advances in neural information processing systems}, 30.

\bibitem[{Vaswani et~al.(2017)Vaswani, Shazeer, Parmar, Uszkoreit, Jones, Gomez, Kaiser, and Polosukhin}]{vaswani2017attention}
Ashish Vaswani, Noam Shazeer, Niki Parmar, Jakob Uszkoreit, Llion Jones, Aidan~N Gomez, \L~ukasz Kaiser, and Illia Polosukhin. 2017.
\newblock \href {https://proceedings.neurips.cc/paper/2017/file/3f5ee243547dee91fbd053c1c4a845aa-Paper.pdf} {Attention is all you need}.
\newblock In \emph{Advances in Neural Information Processing Systems}, volume~30. Curran Associates, Inc.

\bibitem[{Weiss et~al.(2018)Weiss, Goldberg, and Yahav}]{weiss-etal-2018-practical}
Gail Weiss, Yoav Goldberg, and Eran Yahav. 2018.
\newblock \href {https://doi.org/10.18653/v1/P18-2117} {On the practical computational power of finite precision {RNN}s for language recognition}.
\newblock In \emph{Proceedings of the 56th Annual Meeting of the Association for Computational Linguistics (Volume 2: Short Papers)}, pages 740--745, Melbourne, Australia. Association for Computational Linguistics.

\bibitem[{White and Cotterell(2022)}]{white-cotterell-2022-equivariant}
Jennifer~C. White and Ryan Cotterell. 2022.
\newblock \href {https://aclanthology.org/2022.coling-1.412} {Equivariant transduction through invariant alignment}.
\newblock In \emph{Proceedings of the 29th International Conference on Computational Linguistics}, pages 4651--4663, Gyeongju, Republic of Korea. International Committee on Computational Linguistics.

\bibitem[{Wiles(1998)}]{wiles1998recurrent}
Paul Rodriguez~Janet Wiles. 1998.
\newblock Recurrent neural networks can learn to implement symbolsensitive counting.
\newblock \emph{Advances in Neural Information Processing Systems}, 10:87.

\bibitem[{Wong and Wang(2007)}]{wong2007generalisation}
Francis~CK Wong and William~SY Wang. 2007.
\newblock Generalisation towards combinatorial productivity in language acquisition by simple recurrent networks.
\newblock In \emph{2007 International Conference on Integration of Knowledge Intensive Multi-Agent Systems}, pages 139--144. IEEE.

\bibitem[{Xu et~al.(2022)Xu, Niethammer, and Raffel}]{xu2022compositional}
Zhenlin Xu, Marc Niethammer, and Colin~A Raffel. 2022.
\newblock Compositional generalization in unsupervised compositional representation learning: A study on disentanglement and emergent language.
\newblock \emph{Advances in Neural Information Processing Systems}, 35:25074--25087.

\bibitem[{Yin et~al.(2022)Yin, Li, Meng, Zhou, and Zhang}]{yin-etal-2022-categorizing}
Yongjing Yin, Yafu Li, Fandong Meng, Jie Zhou, and Yue Zhang. 2022.
\newblock \href {https://aclanthology.org/2022.coling-1.464} {Categorizing semantic representations for neural machine translation}.
\newblock In \emph{Proceedings of the 29th International Conference on Computational Linguistics}, pages 5227--5239, Gyeongju, Republic of Korea. International Committee on Computational Linguistics.

\bibitem[{Yin et~al.(2023)Yin, Zeng, Li, Meng, Zhou, and Zhang}]{yin-etal-2023-consistency}
Yongjing Yin, Jiali Zeng, Yafu Li, Fandong Meng, Jie Zhou, and Yue Zhang. 2023.
\newblock \href {https://doi.org/10.18653/v1/2023.acl-long.72} {Consistency regularization training for compositional generalization}.
\newblock In \emph{Proceedings of the 61st Annual Meeting of the Association for Computational Linguistics (Volume 1: Long Papers)}, pages 1294--1308, Toronto, Canada. Association for Computational Linguistics.

\bibitem[{Zheng and Lapata(2021)}]{zheng-lapata-2021-compositional-generalization}
Hao Zheng and Mirella Lapata. 2021.
\newblock \href {https://doi.org/10.18653/v1/2021.findings-emnlp.88} {Compositional generalization via semantic tagging}.
\newblock In \emph{Findings of the Association for Computational Linguistics: EMNLP 2021}, pages 1022--1032, Punta Cana, Dominican Republic. Association for Computational Linguistics.

\bibitem[{Zheng and Lapata(2022)}]{zheng-lapata-2022-disentangled}
Hao Zheng and Mirella Lapata. 2022.
\newblock \href {https://doi.org/10.18653/v1/2022.acl-long.293} {Disentangled sequence to sequence learning for compositional generalization}.
\newblock In \emph{Proceedings of the 60th Annual Meeting of the Association for Computational Linguistics (Volume 1: Long Papers)}, pages 4256--4268, Dublin, Ireland. Association for Computational Linguistics.

\bibitem[{Zhou et~al.(2023)Zhou, Jiang, and Bansal}]{zhou-jiang-2023-datafactor}
Xiang Zhou, Yichen Jiang, and Mohit Bansal. 2023.
\newblock Data factors for better compositional generalization.
\newblock In \emph{Proceedings of the 2023 Conference on Empirical Methods in Natural Language Processing}, Abu Dhabi, United Arab Emirates. Association for Computational Linguistics.

\end{thebibliography}

\appendix

\section*{Appendix}
\label{sec:appendix}
\section{Structure-oriented Vector Quantization}
\paragraph{Notations.}
We denote the source sequence as $[x_1 . . . x_N]$ and the target sequence as $[y_1 . . . y_M]$.
The framework consists of an encoder with word embeddings $E_x\in\mathbb{R}^{N_x\times D}$ and an autoregressive decoder with word embeddings $E_y\in\mathbb{R}^{N_y\times D}$, where $N_x$ and $N_y$ are the number of tokens in vocabulary and $D$ is the dimension of the embeddings.
For quantizing $E_x$ and $E_y$, we define two codebooks $Z_x\in\mathbb{R}^{N_x\times D_z}$ and $Z_{y}\in\mathbb{R}^{N_y\times D_z}$, where $N_x$ and $N_y$ are the number of codes and $D_z$ is the dimension of the code embedding.
\subsection{Generalized Brown Clustering Objective.}
\label{appendix_method:elbo_deduction}
Here, we present the details of the Brown Clustering~\cite{brown1992class}., the generalized Brown Clustering objective~\cite{stratos-2019-mutual}, and its Evidence Lower Bound (ELBO)~\eqnref{eqn:elbo_mmi} (with proofs) discussed in~\secref{ssec:method_vq}.

\paragraph{Brown Clustering.}
Brown Clustering is an unsupervised word clustering algorithm that was proposed and popularized before the trend of neural networks.
Brown clustering divides a vocabulary $V$ into $m$ clusters by maximizing the mutual information (MMI) between the clusters $(Z_X, Z_Y)$ of a random bigram $(X,Y)$ in a corpus of N words $(x_1 . . . x_N )$.
The algorithm assumes a uniform distribution over consecutive word pairs $(x_{i-1}, x_i)$ and optimizes the MMI objective:
\begin{equation}
\label{eqn:brown_clustering}
\begin{split}
\max_{Z:V\rightarrow[m]} &= \sum_{z,z'} \frac{\#(z, z')}{N}\mathrm{log}(\frac{\#(z, z')N}{\#(z)\#(z)})\\
\end{split}
\end{equation}
where $\#(z, z')$ denotes the number of occurrences of the cluster pair $(z, z')$ for any bigram in $(x_1 . . . x_N )$.
This objective is intractable, so~\cite{brown1992class} proposed a greedy algorithm that (1) initializes the clusters by assigning each word to a distinct class and then (2) merges the pair of classes that leads to the least loss in the average mutual information for a total of $|V|-m$ times.

This original algorithm failed on vocabularies larger than 5000 words. To the remedy, the authors instead (1) initialize $m$ classes that each contain one of the $m$ most frequent words and (2) repeatedly merge a pair of clusters that yields the smallest decrease in mutual information.
However, this heuristic requires nontrivial combinatorial optimization and is difficult to scale and generalize.

\paragraph{The Generalized MMI Objective.}
\citet{stratos-2019-mutual} generalized the Brown Clustering to the setting by maximizing the mutual information between the posterior clustering probability $q(z|x)$ and prior $p(z|\hat{x})$, where $x$ is a random token from a sentence and $\hat{x}$ is its surrounding context.
Since $p$ and $q$ are conditionally independent given $(x, \hat{x})$, we have $p(z, z'|x,\hat{x})=q(z|x)p(z'|\hat{x})$.
Thus, the mutual information between $p$ and $q$ given a single sentence $(x, \hat{x})$ is:
\begin{equation}
\label{eqn:generalized_brown_clustering}
\begin{split}
J_{x,\hat{x}} &= \sum_{z,z'} q(z|x) p(z'|\hat{x}) \mathrm{log} \frac{q(z|x) p(z'|\hat{x})}{q(z)p(z')}\\
\end{split}
\end{equation}
The author then maximizes $\mathop{\mathbb{E}}_{x,\hat{x}\sim D} [J_{x,\hat{x}}]$. 
This objective becomes the exact Brown Clustering in~\eqnref{eqn:brown_clustering} if (1) $\hat{x}$ is a random context token; (2) $p$ and $q$ are tied and are hard clustering instead of probabilistic soft clustering.

\subsection{A Variational Lower Bound of MMI.}
\citet{stratos-2019-mutual} further improved the generalized Brown Clustering and derived a novel objective as the difference between the entropy of the cluster distribution $H(Z)$ and the cross entropy between $q$ and $p$:
\begin{equation}
\label{eqn:generalized_brown_clustering_elbo}
\begin{split}
J^{var} &= H(Z) - H(q,p)\\
\end{split}
\end{equation}
where $H(q,p)$ is the cross entropy between $q$ and $p$ over samples (\eqnref{eqn:h_q_p}).
Intuitively, minimizing the cross entropy between $q$ and $p$ can improve their mutual information.
Maximizing $H(Z)$ encourages the equal occurrence of each cluster and can prevent the trivial solution that assigns all tokens $x$ in the corpus into the same cluster.

The objective in~\eqnref{eqn:generalized_brown_clustering_elbo} can also be seen as the lower bound of the mutual information between two random variables $I(\hat{X}, Z)$, where $\hat{X}$ is the context of a token $X$ and $Z$ is its cluster inferred from $q(z|x)$. This lower bound is shown in~\citet{mcallester2018information} and we replicate it below.

First, since $I(\hat{X}, Z)=H(Z)-H(Z|\hat{X})$ and by definition we have the conditional entropy:
\begin{equation}
\label{eqn:h_z_x}
\begin{split}
&H(Z|\hat{X})\\
=& -\sum_{\hat{x},z} p(\hat{x}, z) \mathrm{log}\frac{p(\hat{x}, z)}{p(\hat{x})} \\
=& \sum_{\hat{x},z} p(\hat{x}, z) \mathrm{log}\frac{p(\hat{x})}{p(\hat{x}, z)} \\
=& \sum_{\hat{x},z} p(\hat{x}, z) \mathrm{log} \frac{1}{\pi(z|\hat{x})} \\
=&\mathop{\mathbb{E}}_{\substack{(\hat{x},x)\sim D\\z\sim q(\cdot|x)}} [\mathrm{log}\frac{1}{\pi(z|\hat{x})}]\\
\end{split}
\end{equation}
where $\pi(z|\hat{x})$ is the ground-truth prior probability of the cluster of a token given its context.
If we introduce a variational distribution $p(z|\hat{x})$ to approximate $\pi(z|\hat{x})$ and further expand~\eqnref{eqn:h_q_p}, we have
\begin{equation}
\label{eqn:upperbound_of_h_z_x}
\begin{split}
&H(q,p) \\
=&\mathop{\mathbb{E}}_{(x,\hat{x})\sim D}\left[-\sum_{z}q(z|x)\mathrm{log}p(z|\hat{x})\right] \\
=&\mathop{\mathbb{E}}_{(x,\hat{x})\sim D}\left[\sum_{z}q(z|x)\mathrm{log}\frac{1}{p(z|\hat{x}})\right] \\
=& \mathop{\mathbb{E}}_{\substack{(\hat{x},x)\sim D\\z\sim q(\cdot|x)}} \left[\mathrm{log}\frac{1}{p(z|\hat{x})}\right] \\
=& \mathop{\mathbb{E}}_{\substack{(\hat{x},x)\sim D\\z\sim q(\cdot|x)}} \left[\mathrm{log}\frac{\pi(z|\hat{x})}{\pi(z|\hat{x})p(z|\hat{x})}\right] \\
=& \mathop{\mathbb{E}}_{\substack{(\hat{x},x)\sim D\\z\sim q(\cdot|x)}} \left[\mathrm{log}\frac{1}{\pi(z|\hat{x})}\right] + \mathop{\mathbb{E}}_{\substack{(\hat{x},x)\sim D\\z\sim q(\cdot|x)}} \left[\mathrm{log}\frac{\pi(z|\hat{x})}{p(z|\hat{x})}\right] \\
=& H(Z|\hat{X}) + D_{\mathrm{KL}}(\pi||p) \\
\end{split}
\end{equation}
where $D_{\mathrm{KL}}$ is Kullback–Leibler divergence. Therefore, we can see that $H(p,q)$ is an upper bound of $H(Z|\hat{X})$, and hence $H(z)-H(q,p)$ is the lower bound of $I(\hat{X},Z)$.
\citet{stratos-2019-mutual} argues that maximizing $I(\hat{X}, Z)$ over the cluster inference distribution $q$ enforces ``predictive coding'' because it pushes the cluster inferred from $q(z|X)$ to be more informative of the context $\hat{X}$. 

\subsection{The Variational Brown Clustering}
\label{appendix_method:reimagined_brown_clustering}
In this work, we propose another MMI objective that marries the original Brown Clustering objective (\eqnref{eqn:brown_clustering}) and the alternative objective (\eqnref{eqn:elbo_mmi}).
First, we redefine the cluster prediction distribution as $p(z|\hat{z})$, where $\hat{z}$ are the quantized codes of all context tokens $\hat{x}$ inferred from $\mathrm{argmax}(q(z|\hat{x}))$.
This differs from the $p(z|\hat{x})$ that predicts the cluster of $x$ directly from its context $\hat{x}$.
Then, instead of maximizing the ELBO of $I(\hat{X}, Z)$, we maximize the ELBO of $I(\hat{Z}, Z)$. 
Next, we derive the lower bound of $I(\hat{Z}, Z)$ in a similar way to~\eqnref{eqn:h_z_x} and~\eqnref{eqn:upperbound_of_h_z_x}.
First, by definition we have:
\begin{equation*}
I(\hat{Z},Z) = H(Z) - H(Z|\hat{Z})
\end{equation*}

Then, we rewrite $H(Z|\hat{Z})$ as:
\begin{equation}
\label{eqn:h_z_zhat}
\begin{split}
&H(Z|\hat{Z})\\
=& -\sum_{\hat{z},z} p(\hat{z}, z) \mathrm{log}\frac{p(\hat{z}, z)}{p(\hat{z})} \\
=& \sum_{\hat{z},z} p(\hat{z}, z) \mathrm{log}\frac{p(\hat{z})}{p(\hat{z}, z)} \\
=& \sum_{\hat{z},z} p(\hat{z}, z) \mathrm{log} \frac{1}{\pi(z|\hat{z})} \\
=& \sum_{\hat{z},z}\sum_{\hat{x},x}p(\hat{x},x)p(\hat{z}, z|\hat{x},x) \mathrm{log} \frac{1}{\pi(z|\hat{z})} \\
=&\mathop{\mathbb{E}}_{\substack{(\hat{x},x)\sim D\\z\sim q(\cdot|x)\\\hat{z}\sim q(\cdot|\hat{x})}} [\mathrm{log}\frac{1}{\pi(z|\hat{z})}]\\
\end{split}
\end{equation}
where $\pi(z|\hat{z})$ is the ground-truth code prediction distribution given its context's codes. If we introduce a variational distribution $p(z|\hat{z})$ and rewrite~\eqnref{eqn:h_q_p} with the newly defined $p(z|\hat{z})$, we have:
\begin{equation*}
\begin{split}
&H(q,p) \\
=&\mathop{\mathbb{E}}_{\substack{(\hat{x},x)\sim D\\\hat{z}\sim q(\cdot|\hat{x})}}\left[-\sum_{z}q(z|x)\mathrm{log}p(z|\hat{z})\right] \\
=&\mathop{\mathbb{E}}_{\substack{(\hat{x},x)\sim D\\\hat{z}\sim q(\cdot|\hat{x})}}\left[\sum_{z}q(z|x)\mathrm{log}\frac{1}{p(z|\hat{x})}\right] \\
=& \mathop{\mathbb{E}}_{\substack{(\hat{x},x)\sim D\\\hat{z}\sim q(\cdot|\hat{x})\\z\sim q(\cdot|x)}} \left[\mathrm{log}\frac{1}{p(z|\hat{z})}\right] \\
=& \mathop{\mathbb{E}}_{\substack{(\hat{x},x)\sim D\\\hat{z}\sim q(\cdot|\hat{x})\\z\sim q(\cdot|x)}} \left[\mathrm{log}\frac{\pi(z|\hat{z})}{\pi(z|\hat{x})p(z|\hat{z})}\right] \\
=& \mathop{\mathbb{E}}_{\substack{(\hat{x},x)\sim D\\\hat{z}\sim q(\cdot|\hat{x})\\z\sim q(\cdot|x)}} \left[\mathrm{log}\frac{1}{\pi(z|\hat{z})}\right] + \mathop{\mathbb{E}}_{\substack{(\hat{x},x)\sim D\\\hat{z}\sim q(\cdot|\hat{x})\\z\sim q(\cdot|x)}} \left[\mathrm{log}\frac{\pi(z|\hat{z})}{p(z|\hat{z})}\right] \\
=& H(Z|\hat{Z}) + D_{\mathrm{KL}}(\pi||p) \\
\end{split}
\end{equation*}
Therefore, $H(q,p)$ is the upper bound of $H(Z|\hat{Z})$, and hence $H(Z) - H(q,p)$ is the lower bound of $I(\hat{Z}, Z)$.

\subsection{Implementation of MMI objectives}
\label{appendixssec:implementation}
Finally, we show our empirical implementation of the MMI objective.
First, we implement $q(z|x)$ as the cosine similarity between the word embedding $e_x$ and code embedding $z$.
We implement $p(z|\hat{x})$ as a separate Transformer network that takes the context\footnote{The context can be either the whole sentence with $x$ masked (bidirectional context) or the preceding words of $x$ only (left context).} of $x$ as the input and predicts the class of $x$.
Following~\citet{stratos-2019-mutual}, we estimate these terms from the training data:
\begin{equation}
\label{eqn:empirical_objectives}
\begin{split}
q'(z) &= \frac{1}{N}\sum_{i=1}^{N}q(z|x_i) \\
H'(Z) &= -\sum_{z}q'(z)\mathrm{log}q'(z) \\
H'(p,q) &= \frac{1}{N}\sum_{i=1}^{N}(-\sum_{z}q(z|x_i)\mathrm{log}p(z|\hat{z_i}))\\
\end{split}
\end{equation}
where $N$ is the number of tokens.
We then add $\alpha(H'(p,q)-H'(Z))$ to the overall loss, where $\alpha$ is a tunable coefficient.

\subsection{Systematic Attention Layer for Decoder}
\label{appendix_ssec:sal_decoder}
In~\secref{ssec:method_sal}, we introduce the Systematic Attention Layer (SAL) for a Transformer encoder.
Here, we introduce SAL for a Transformer decoder with encoder-decoder cross attention.
we can use the quantized embedding $z$ as the queries and keys in computing the self-attention weights but use the word embedding $x$ as the queries and keys for cross attention.
This is because we find that long-term, cross-sentence relationships cannot be determined by words' syntactic functions only (\secref{ssec:discussion_compositionality}):
\begin{equation*}
\begin{split}
z_{l} &+= \mathrm{SelfAttn}(q=z_{l-1},k=z_{l-1},v=z_{l-1}) \\
z_{l} &+= \mathrm{CrossAttn}(q=z_l,k=z^x_L,v=z^x_L) \\
y_{l} &+= \mathrm{SelfAttn}(q=z_{l-1},k=z_{l-1},v=y_{l-1}) \\
y_{l} &+= \mathrm{CrossAttn}(q=y_l,k=x_L,v=x_L) \\
\end{split}
\end{equation*}
where $y_0$ and $z_0$ are the non-contextualized word embeddings and their quantized code embeddings respectively,
$x_L$ and $z^x_L$ are the encoder's final layer outputs (as shown in~\figref{fig:sal}).
In the equation above, the cross-attention that computes $y_l$ (last row) is the same as a regular Transformer layer.
After the final layer, we project $z_L$ and supervise it to predict the code of the next token, same as how we project $x_L$ to predict the next token.

\subsection{Information Bottleneck Interpretation}
\citet{tishby2015deep} state that to have a generalizable deep neural network, it is necessary to optimize the Information Bottleneck (IB) tradeoff between compression and prediction for each layer. 
It is equivalent to minimizing the Lagrangian $I(X_{l-1},X_l) - \beta I(X_l,Y)$, where $X_l$ is a mapping (e.g., representation produced by the $l$-th layer in a neural net) of $X$ and $I(X_l,Y)$ is the mutual information between $X_l$ and the label $Y$. 
This objective suggests that we need to find the most concise
representation $X_l$ that is also sufficient to encode $I(X_l,Y)$ at each layer $l$.
On the one hand, minimizing $I(X_{l-1},X_l)$ can prevent redundant information from flowing to the next layer, which could be exploited to establish some spurious correlations between $X_l$ and $Y$.
On the other hand, maximizing $I(X_l,Y)$ ensures that sufficient information is encoded in layer $l$ to enables the final prediction of $Y$.

We argue that SoVQ and SRL can implicitly minimize $I(X_{l-1},X_l)$ for $l=0,1...L$:
(1) SoVQ clusters the $N$ word embeddings $X_0$ around $K$ code embeddings ($N>Z$\footnote{For example, we only use 16 codes ($K$=16) to quantize 40356 subword embeddings for the WMT17 En-De tasks.}), thus achieving a lower $H(X_0)$ compared to an unrestricted embedding space and minimizing $I(X,X_0)$;
(2) SRL, on the other hand, clusters the layer's outputs computed from word embeddings around the layer's outputs computed from code embeddings. 
It thus reduces $H(X_l)$ to minimize $I(X_{l-1},X_l)$ for $l=1...L$.
Therefore, from the standpoint of information theory, SoVQ and SRL impose information bottlenecks on the embedding layer and every attention layer to improve the generalization of the entire network.

\begin{figure*}[t]
\centering
\subfloat[Attention maps encoding ``\textit{walk around left}'', trained on original \addjump.]{\label{subfig:1x_attn_maps_walkaroundleft_2} \includegraphics[width=0.23\textwidth]{figures/x1_attn_maps_walkaroundleft.pdf}}
\hfill
\subfloat[Attention maps encoding ``\textit{look around left}'', trained on original \addjump.]{\label{subfig:1x_attn_maps_lookaroundleft_2} \includegraphics[width=0.23\textwidth]{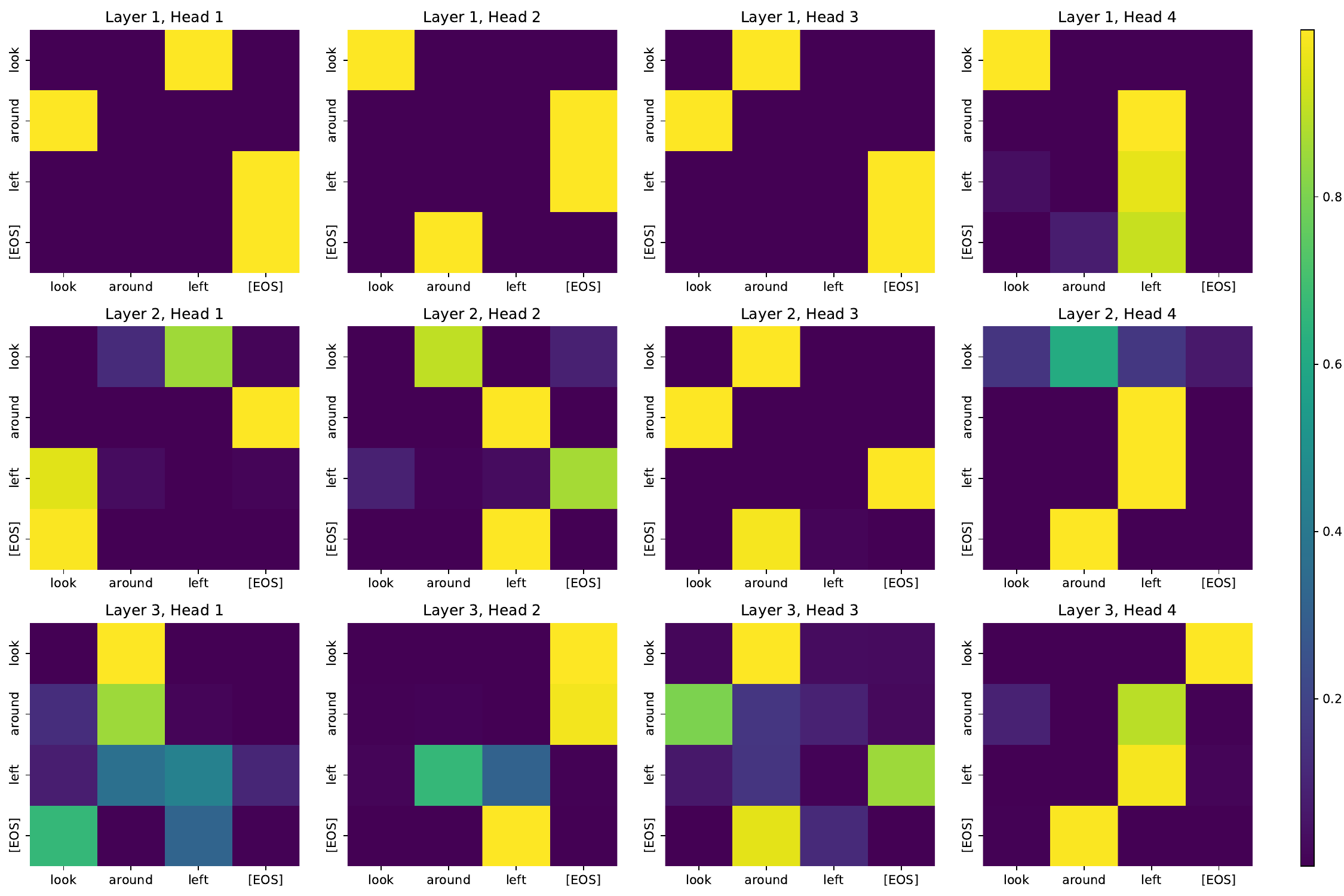}}
\hfill
\subfloat[Attention maps encoding ``\textit{run around left}'', trained on original \addjump.]{\label{subfig:1x_attn_maps_runaroundleft_2} \includegraphics[width=0.23\textwidth]{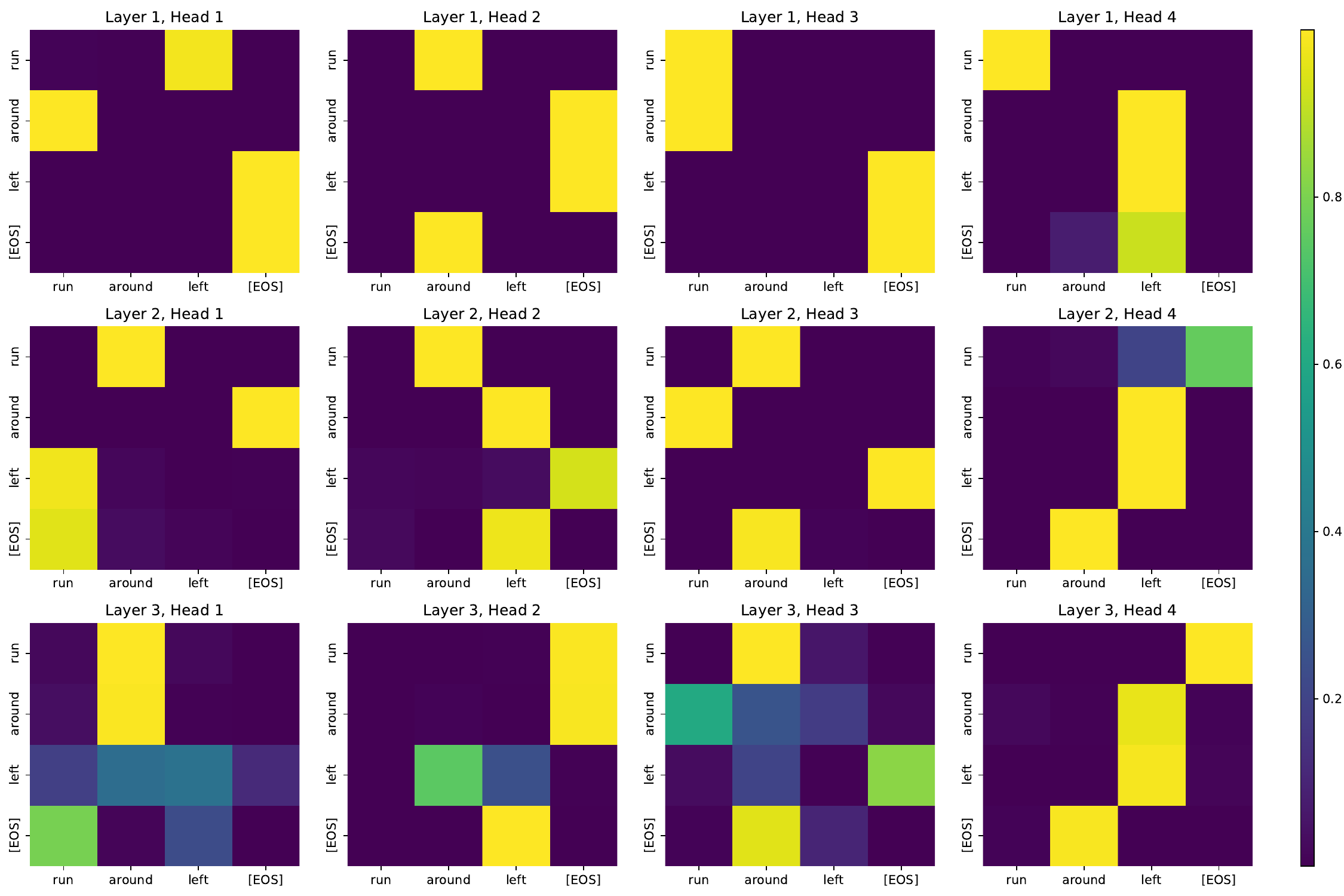}}
\hfill
\subfloat[Attention maps encoding ``\textit{jump around left}'', trained on original \addjump.]{\label{subfig:1x_attn_maps_jumparoundleft_2} \includegraphics[width=0.23\textwidth]{figures/x1_attn_maps_jumparoundleft_v2_thicker.pdf}}
  \vspace{0.5em}
\subfloat[Attention maps encoding ``\textit{walk around left}'', trained on 20x augmented \addjump.]{\label{subfig:20x_attn_maps_walkaroundleft_2} \includegraphics[width=0.23\textwidth]{figures/x20_attn_maps_walkaroundleft.pdf}}
\hfill
\subfloat[Attention maps encoding ``\textit{look around left}'', trained on 20x augmented \addjump.]{\label{subfig:20x_attn_maps_lookaroundleft_2} \includegraphics[width=0.23\textwidth]{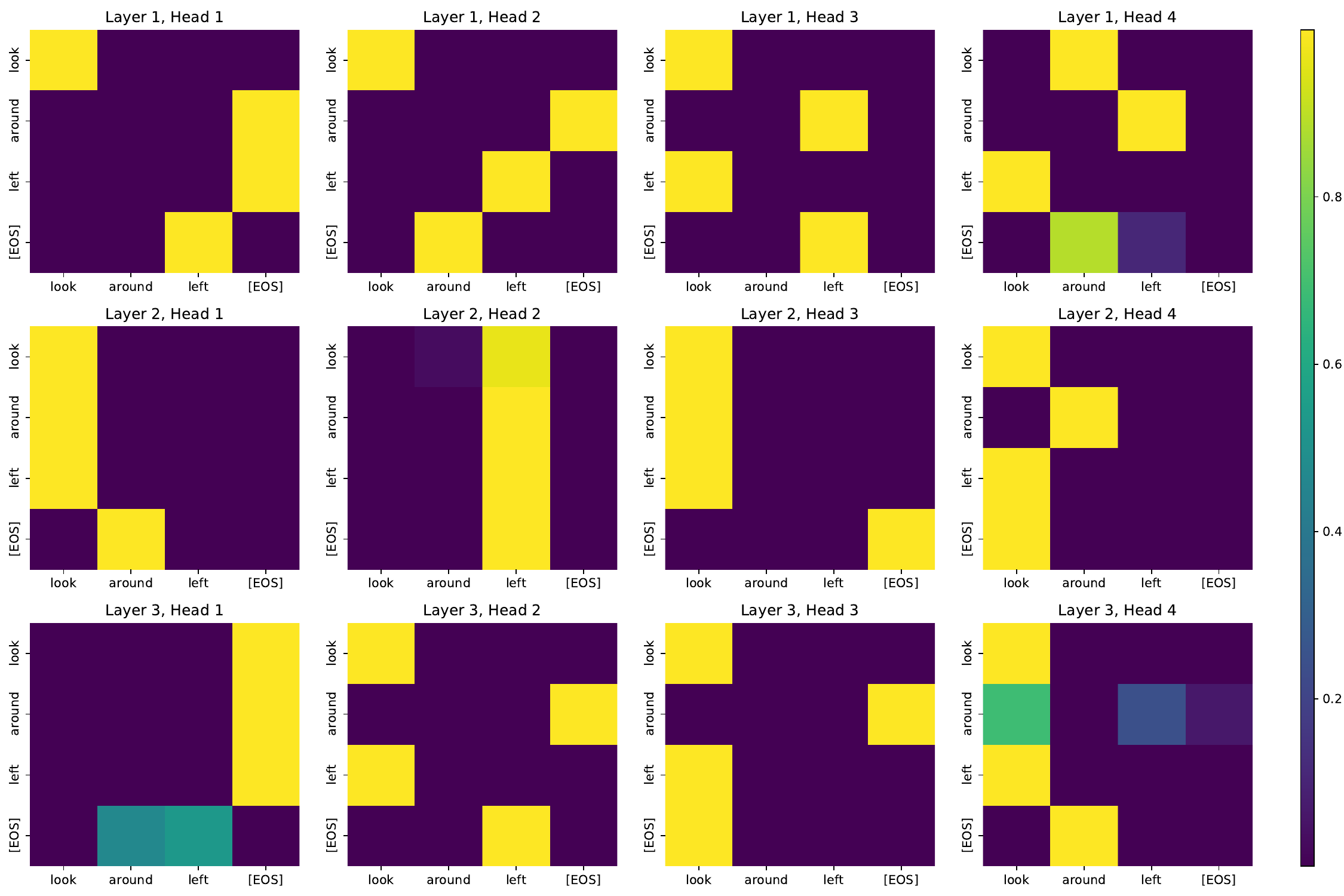}}
\hfill
\subfloat[Attention maps encoding ``\textit{run  around left}'', trained on 20x augmented \addjump.]{\label{subfig:20x_attn_maps_runaroundleft_2} \includegraphics[width=0.23\textwidth]{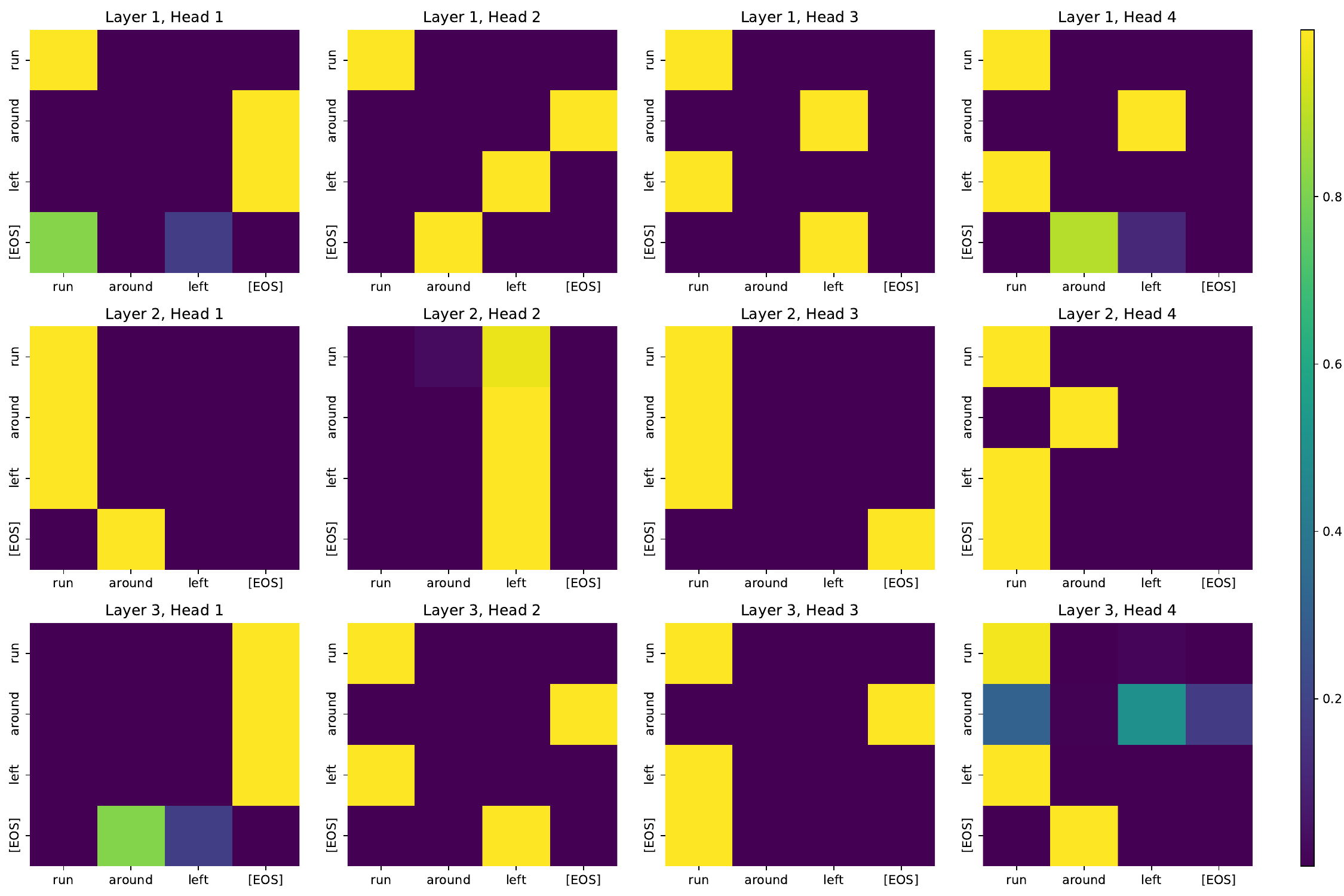}}
\hfill
\subfloat[Attention maps encoding ``\textit{jump around left}'', trained on 20x augmented \addjump.]{\label{subfig:20x_attn_maps_jumparoundleft_2} \includegraphics[width=0.23\textwidth]{figures/x20_attn_maps_jumparoundleft.pdf}}
\caption{Attention maps encoding training examples ``\textit{walk around left}'', ``\textit{look around left}'', ``\textit{run around left}'' and a test example ``\textit{jump around left}'' from the Transformers trained on the original SCAN \addjump training set (a, b, c, d), and 20x augmented training set~\cite{zhou-jiang-2023-datafactor} (e, f, g, h).
We highlight the attention patterns in (d) that differ from (a), (b), (c) in red boxes.
When trained on 20x augmented training set, the model encodes the two examples (c and d) with highly similar attention maps across all layers and heads. 
}
\label{fig:1x_20x_lookrunwalkjump_attn_maps}
\end{figure*}

\section{Experiments}
\subsection{Experimental Setup}
\label{appendix_ssec:experimental_setup}
\paragraph{Semantic parsing experiments.}
We use a 3-layer Transformer encoder and a 3-layer Transformer decoder with 4 heads per layer, a hidden size of 256, and a feedforward size of 512. 
We share input and output embeddings of the decoder. 
We optimize the model using Adam~\cite{kingma15adam}, with $\beta_1$ = 0.1, $\beta_2$ = 0.98. 
All models are trained for 100,000 steps and we choose the best checkpoint on validation set for evaluation.
On SCAN tasks where the original vocab size is small (17 for source and 10 for target, including special tokens), we try source codebook sizes [4,6,8] and target codebook sizes [3,4,5], and end up using 6 codes for quantizing source tokens and 4 codes for quantizing target tokens.
On COGS with 747 source tokens and 687 target tokens, we try source and target codebook sizes [8,16,32], and end up using 32 codes for quantizing source tokens and 16 codes for quantizing target tokens.

\paragraph{Machine translation experiments.}
We use a 6-layer Transformer encoder and a 6-layer Transformer decoder with 8 heads per layer, a hidden size of 512, and a feedforward size of 1024.
It has the same size as the Transformer used in~\citet{yin-etal-2023-consistency}.
We share input and output embeddings of the decoder. 
The model parameters are optimized
by Adam~\cite{kingma15adam}, with $\beta_1$ = 0.1, $\beta_2$= 0.98. 
All models are trained for 1 million steps on CoGnition and 2 million steps on WMT training sets.
We then choose the best checkpoint on the validation set for evaluation.
During decoding, we use a beam size of 5 and a maximum generation length of $ax+b$ where $a$=1.2 and $b$=10.
On CoGnition with 2004 source English tokens and 5500 target Chinese tokens (including special tokens), we try source and target codebook sizes [16,32,64,128], and end up using 64 codes for quantizing source tokens and 32 codes for quantizing target tokens.
On WMT tasks, we follow the tokenization and preprocessing steps in fairseq\footnote{\url{https://github.com/facebookresearch/fairseq}}, and use 16 codes each for quantizing source tokens and target tokens.
\begin{figure*}[t]
\centering
\subfloat[Source embeddings of the Baseline.]{\label{subfig:cogs_src_baseline} \includegraphics[width=0.32\textwidth]{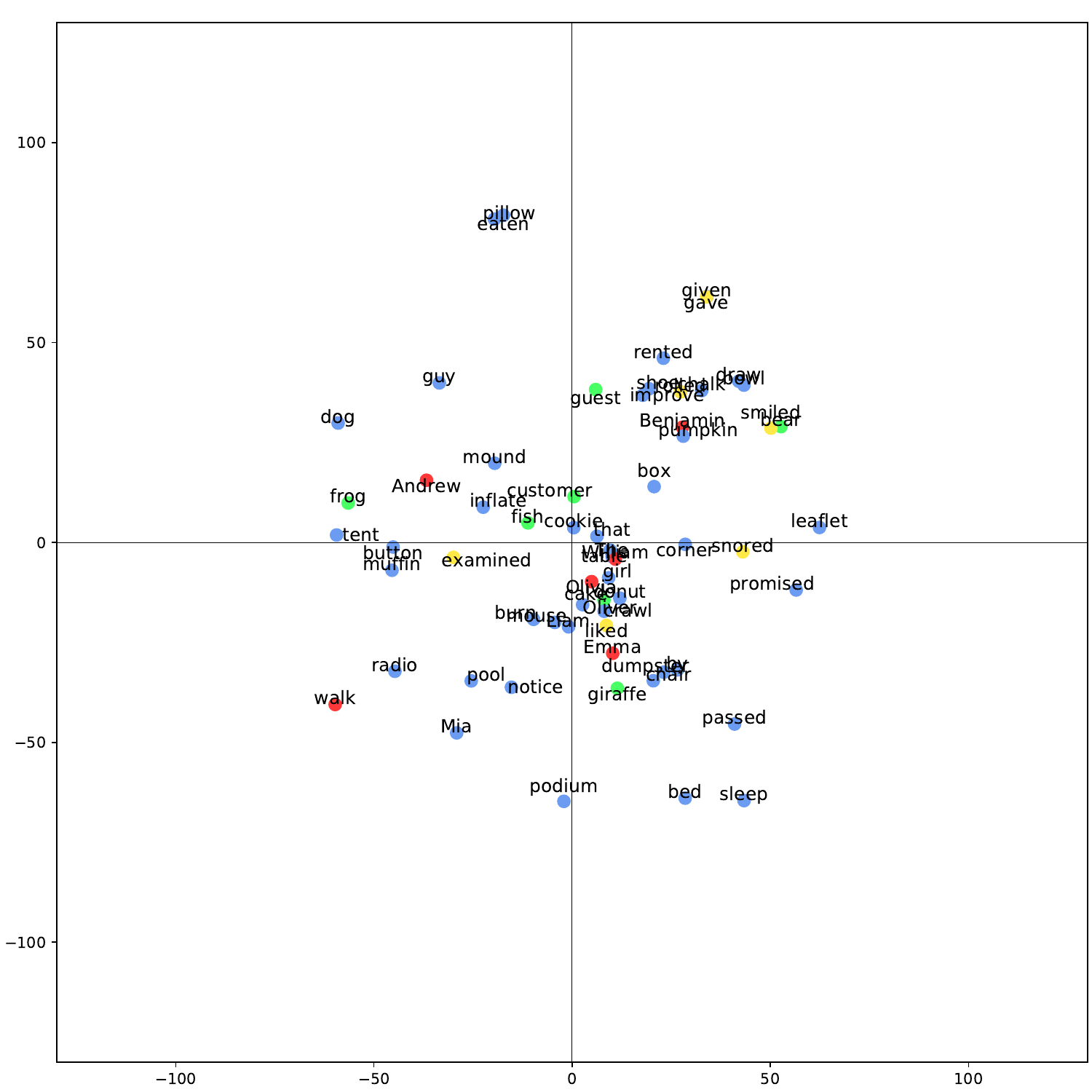}}
\hfill
\subfloat[Source embeddings with VQ.]{\label{subfig:cogs_src_vq} \includegraphics[width=0.32\textwidth]{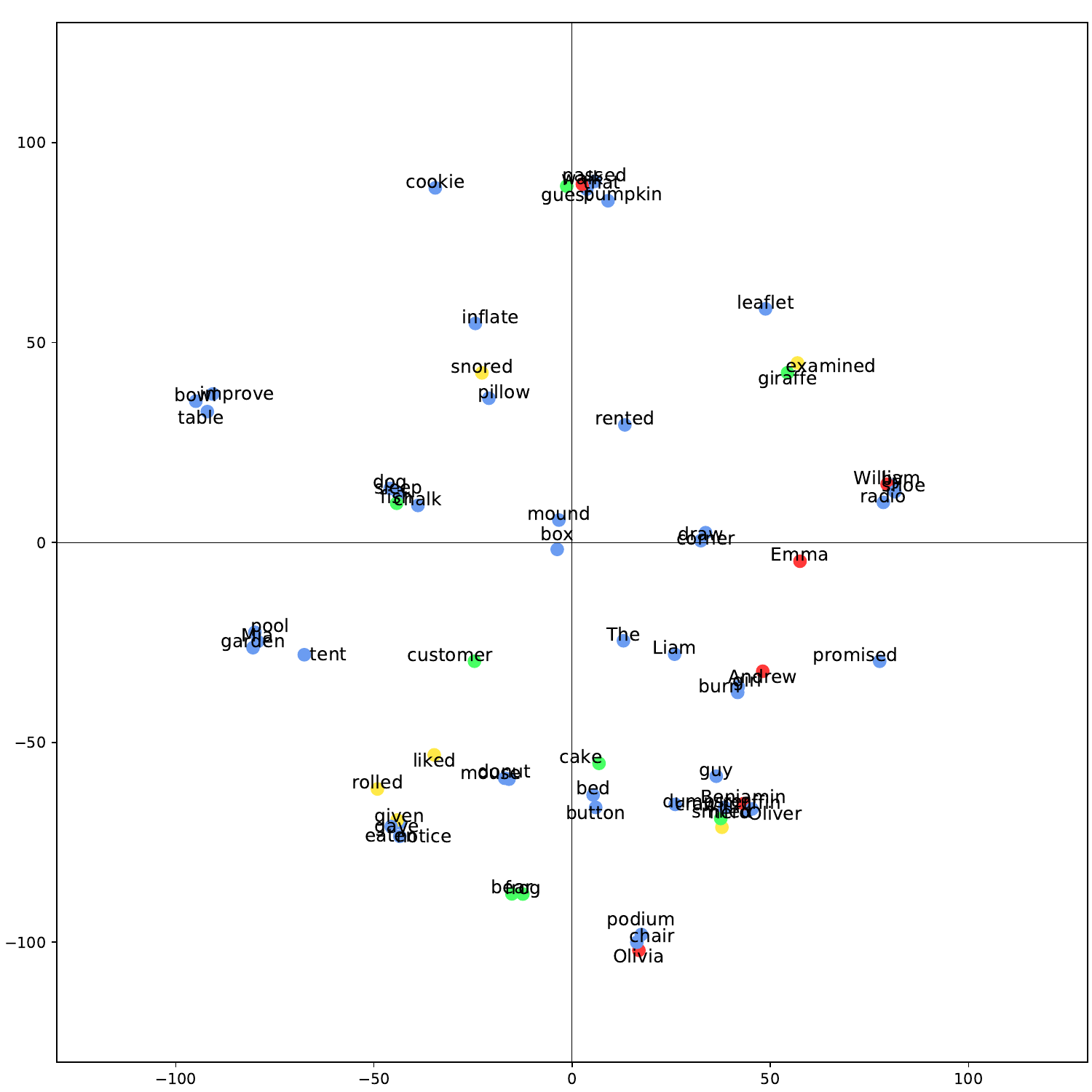}}
\hfill
\subfloat[Source embeddings with SoVQ.]{\label{subfig:cogs_src_sovq} \includegraphics[width=0.32\textwidth]{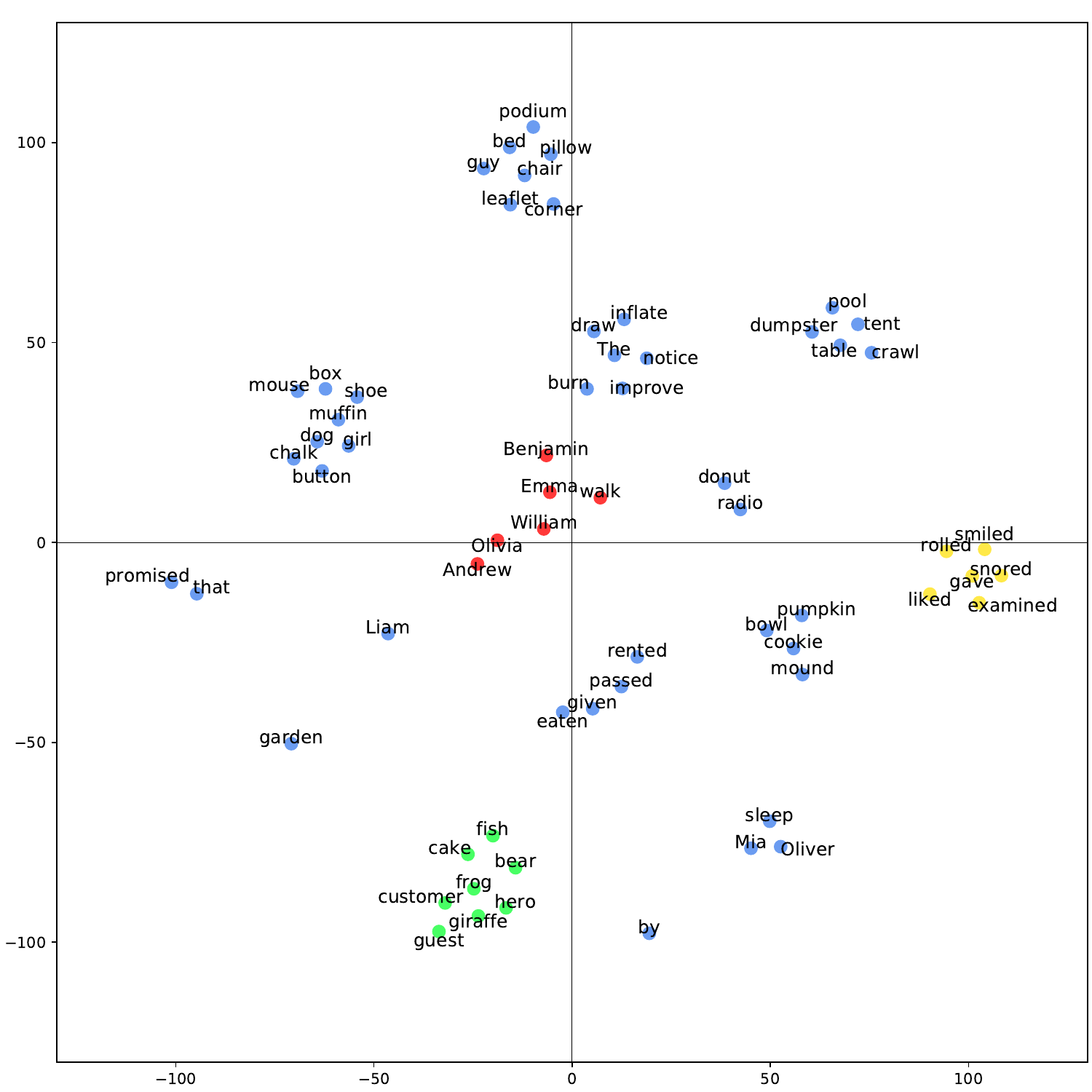}}%
\caption{T-SNE of embeddings learned on \textsc{COGS}~\cite{kim-linzen-2020-cogs}, with 32 clusters used in VQ.}
\label{fig:tsne_cogs}
\end{figure*}

\section{Additional Analyses}

\subsection{Visualizing COGS Embedding Space}
\label{appendixssec:cogs_embed_visualization}
In~\figref{fig:tsne_cogs}, we show the t-SNE of the source embedding matrices learned on COGS.
Compared to SCAN, COGS has a much larger vocabulary (748 VS 13) and more diverse syntactic structures.
Again, in~\figref{subfig:cogs_src_sovq} we show that SoVQ can cluster the word embeddings based on their syntactic functions.
For example, the ``red cluster'' (in the middle of 2-d t-SNE space) is mostly made of names like ``Andrew'' and ``Olivia''.
The ``yellow cluster'' on the right side is comprised of verbs in their past participle forms (e.g., ``smiled, rolled''). 
The ``green'' cluster at the bottom includes animals like ``fish'', ``bear'', and ``giraffe''.
In~\tabref{table:cogs_closest_words}, we show 10 words sampled from the source vocabulary and their closest neighbors in the 2-d t-SNE space. 
One interesting finding is that the word `can' is clustered with nouns like `block', `bee', and `ring'. This is because in the COGS training set, `can' is only used as a noun and never used as a modal verb.

The embedding space learned by the Transformer baseline, on the other hand, does not demonstrate any patterns that can connect the distribution and the syntactic role of a word~\figref{subfig:cogs_src_baseline}.
We can observe that sometimes words that are different tenses of the same verb (e.g., ``given'' and ``gave'' in the red circle) or have connections in their semantics (e.g., ``sleep'' and ``bed'' in the blue circle) are clustered together.
The Transformer with vanilla Vector Quantization, although learns a more cluster-separated embedding space~\figref{subfig:cogs_src_vq}, also does not demonstrate any noticeable similarities of words within a cluster. 

Overall, based on the difference in how words are clustered in the embedding space, we can state that Structural-oriented Vector Quantization (SoVQ) can effectively cluster words based on their syntactic functions.

\begin{table*}[t!]
\centering
\begin{small}
\begin{tabular}[t]{l|l}
\toprule
 \multicolumn{1}{c}{Source Word} & \multicolumn{1}{c}{Closest Words} \\
\midrule
William&Sophia, Riley, Carter, Nora, Madison, John, Jack, Lillian, Sebastian, Christopher\\ 
draw&float, love, scream, double, see, shark, help, roll, frown, worship\\ 
ate&gave, wanted, noticed, laughed, smirked, hoped, talked, napped, doubled\\ 
bear&manager, writer, cow, warrior, governor, crocodile, bicycle, boulder, bag, cloud\\ 
cookie&piano, rug, car, banana, melon, bench, bottle, bible, storage, turtle\\ 
improve&redden, poke, pierce, discover, throw, toss, slide, freeze, disintegrate, Paula\\ 
bed&taxi, pot, sphere, cot, couch, bunker, backpack, glacier, vehicle, bin\\ 
preferred&jogged, smiled, sketched, craved, touched, gasped, yearned, supported, saw, crumple\\ 
teacher&strawberry, raisin, beast, soap, monster, clock, rose, child, lawyer, chief\\ 
can&block, bee, ring, blender, tripod, seat, jacket, dog, donkey, beer\\ 
table&stage, speaker, desk, barrel, boat, trunk, house, room, stand\\ 
\bottomrule
\end{tabular}
\caption{Words sampled from the source vocabulary of COGS and their closest words in the 2-d t-SNE space.
}
\label{table:cogs_closest_words}
\end{small}
\end{table*}

\begin{table*}[t!]
\centering
\begin{small}
\begin{tabular}[t]{p{7.5cm}|p{7.5cm}}
\toprule
 \multicolumn{1}{c}{Sentence 1} & \multicolumn{1}{c}{Sentence 2} \\
\midrule
She chose another child he liked .&She chose another clown he liked .\\ 
She also found every small girl .&She also met each small clown .\\ 
He took every large clown .&He caught every special clown !\\ 
He heard each large girl .&He saw each small girl .\\ 
He heard every silly girl .&He found every dirty girl .\\ 
The man took every large clown .&The cat caught every special clown .\\ 
He stopped every girl .&He met each girl .\\ 
She saw the silly clown .&She watched the dirty girl .\\ 
She met each small clown .&She watched every large girl .\\ 
She invited the large clown .&She saw the special child .\\ 
The smart doctor he liked was very proud .&The dirty dog he liked was extremely excited .\\ 
He invited each small boyfriend on the floor .&He found every large boyfriend on the floor .\\ 
He left every small girl he liked .&He caught each large child he liked .\\ 
Taylor met the special girl .&Taylor heard the small clown .\\ 
He visited the large child he liked .&He heard the large girl he liked .\\ 
I chose another clown he liked .&I chose another child he liked .\\ 
She visited each silly boyfriend on the floor .&She found every dirty boyfriend on the floor .\\ 
He hated the large clown on the floor .&He applied to the large doctor on the floor .\\ 
Taylor hated the small building .&Taylor hated the large building .\\ 
Any red doctor at the store jumped back .&Any small farmer at the store pushed him .\\ 
She woke any large building on the floor .&She woke any small building on the floor .\\ 
I chose each large boyfriend on the floor .&I chose each small car on the floor .\\ 
His neighbors heard the special clown on the floor .&His friend heard the small clown on the floor .\\ 
He looked under any small chair on the floor .&He looked under any small chair on the floor .\\ 
There was a hurricane headed towards each small doctor .&There was a hurricane headed towards every small doctor .\\ 
Taylor hated every silly bee .&Taylor went to each empty bee .\\ 
Taylor was sad about every small building .&Taylor was excited about every small building .\\ 
She had everything taken care of except every small chair .&She had everything taken care of except each large apartment .\\ 
Taylor visited each silly boyfriend on the floor .&Taylor found every silly boyfriend on the floor .\\ 
Taylor lost another small girl on the floor .&Taylor lost another small clown on the floor .\\ 
Taylor watched every dirty girl he liked .&Taylor saw every silly clown he liked .\\ 
Taylor chose each small girl he liked .&Taylor caught each large child he liked .\\ 
Taylor woke each large clown for school .&Taylor woke every small clown for school .\\ 
Taylor visited the small child he liked .&Taylor heard the large girl he liked .\\ 
Taylor caught any large car on the floor .&Taylor took any small boyfriend on the floor .\\ 
Taylor found every large child on the floor .&Taylor heard each small clown on the floor .\\ 
Taylor watched each dirty clown on the floor .&Taylor heard every dirty clown on the floor .\\ 
He smiled and gave each car a free popcorn .&He smiled and gave every boyfriend a free popcorn .\\ 
She invited all the girls except any small building on the floor .&She invited all the girls except any small farm on the floor .\\ 
Taylor hated every silly bee he liked .&Taylor hated each dirty bee he liked .\\ 
Taylor hated any large bee on the floor .&Taylor hated any small bee on the floor .\\ 
I woke the silly child on the floor up to give him a sandwich .&I woke the dirty clown on the floor up to give him a sandwich .\\ 
Another smart doctor he liked got so bad that she could n't stand it .&Another smart doctor he liked got so bad that she could n't stand it .\\ 
Except each empty apartment he liked , she took it out to show to a friend .&Except every empty airplane he liked , she took it out to show to a friend .\\ 
When i got home , i did all my homework except each empty apartment he liked .&When i got home , i did all my homework except every empty airplane he liked .\\ 
When i got home , i did all my homework except the quiet farm he liked .&When i got home , i did all my homework except the empty building he liked .\\ 
Taylor stayed inside the small building on the floor , even though a storm was coming .&Taylor stayed inside the large farm on the floor , even though a storm was coming .\\ 
As soon as i was about to take a bath , i saw a light inside any empty car he liked .&As soon as i was about to take a bath , i saw a light inside any quiet car he liked .\\ 
\bottomrule
\end{tabular}
\caption{Source sentences collected from the CoGnition~\cite{li-etal-2021-compositional} compositional generalization test set.
The sentences within each pair are quantized into the same sequence of clusters by our \modelname. 
}
\label{table:same_code_seq_examples}
\end{small}
\end{table*}

\subsection{Analyzing the Attention Pattern of SRL.}
We show the example pairs used in~\secref{ssec:analysis_circuits}.
We collect a total of 48 pairs of source sentences from the test set.
The source sentences within each pair are quantized into the same sequence of clusters (e.g., ``\textit{he stopped every girl.}'' and ``\textit{he found each child.}'') by our \modelname.

\end{document}